\documentclass[a4paper]{article}

\usepackage{xcolor, soul}
\usepackage{graphicx}
\usepackage{spverbatim}
\usepackage[bookmarks=false]{hyperref}
\usepackage{authblk}
\usepackage{adjustbox}
\usepackage[square,numbers]{natbib}
\usepackage{booktabs}

\begin{document}

\title{Concerns and Values in Human-Robot Interactions: A Focus on Social Robotics}

\author[*]{Giulio Antonio Abbo}
\author[*]{Tony Belpaeme}
\author[**]{Micol Spitale}

\affil[*]{\textit{IDLab-AIRO}, \textit{Ghent University -- imec}, Ghent, Belgium}
\affil[**]{\textit{DEIB}, \textit{Politecnico di Milano}, Milano, Italy}
\affil[ ]{\texttt{giulioantonio.abbo@ugent.be}}

\date{\scriptsize January 2025 -- To appear in the International Journal of Social Robotics}

\maketitle

\begin{abstract}
Robots, as AI with physical instantiation, inhabit our social and physical world, where their actions have both social and physical consequences, posing challenges for researchers when designing social robots. This study starts with a scoping review to identify discussions and potential concerns arising from interactions with robotic systems in the context of healthcare, education, and private homes. Two focus groups of technology ethics experts then validated a comprehensive list of key topics and values in human-robot interaction (HRI) literature in these contexts. These insights were integrated into the HRI Value Compass web tool, to help HRI researchers identify these values in robot design. The tool was evaluated in a pilot study. This work benefits the HRI community by highlighting key concerns in human-robot interactions and providing an instrument to help researchers design robots that align with human values, ensuring future robotic systems adhere to these values in social applications.
\end{abstract}

\begin{figure*}[h!]
  \centering
  \begin{adjustbox}{width=1.1\textwidth,center=\textwidth}
  \includegraphics[width=\textwidth]{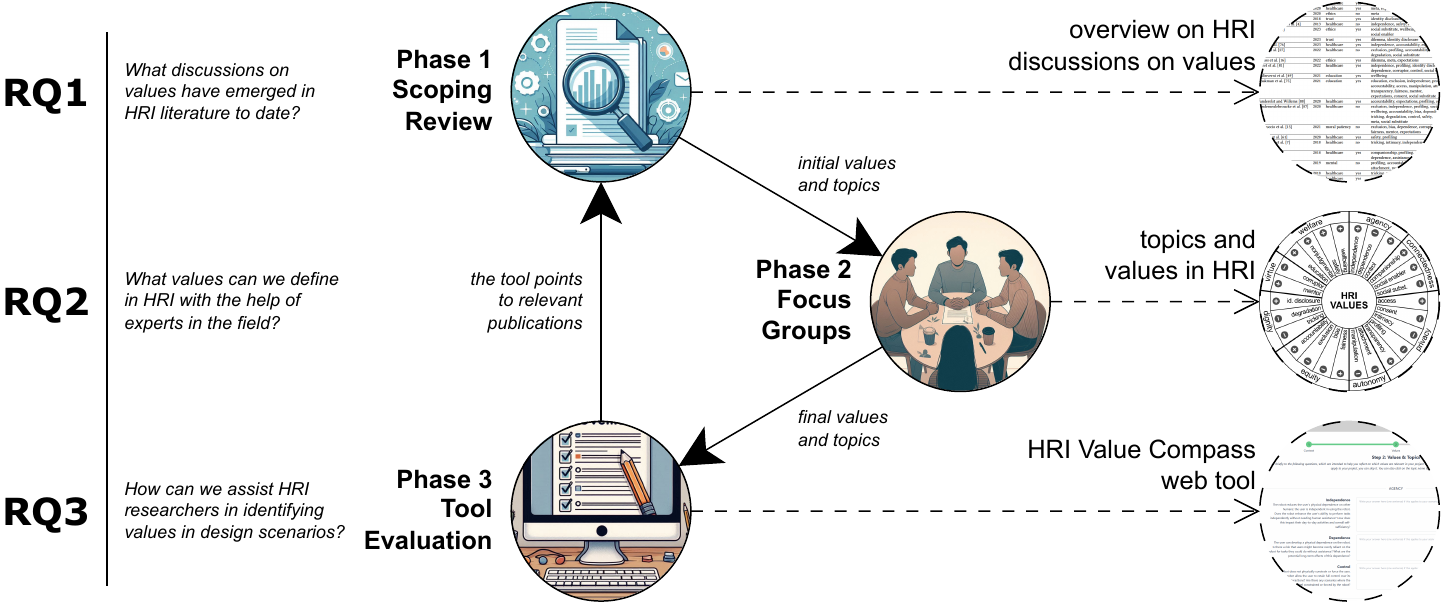}
  \end{adjustbox}
  \caption{The research questions (left), the design process (centre), and the contributions (right).}
  \label{fig:steps}
\end{figure*}

\section{Introduction}

Recent advancements in artificial intelligence and robotics are pushing robots from research laboratories into public spaces~\cite{mintrom2022robots}, healthcare facilities~\cite{kyrarini2021survey} and schools~\cite{benitti2012exploring}, as well as private homes.
As robots start exhibiting nuanced, multifaceted behaviours, by inhabiting high-stakes environments where their actions have profound social and physical consequences, robots become more and more social agents~\cite{bartneck2024human}.
As such, they transcend their nature of complex tools and start to pose new ethical challenges, impacting the humans involved in immediate and long-term interactions.

Human-robot interaction (HRI) researchers must anticipate and address these ethical concerns.
This involves understanding the potential risks and benefits associated with robot interactions and striving to embed ethical principles and values into the design of robotic systems that align with societal norms and values.
In social psychology, seminal works such as Schwartz's theory of basic human values~\cite{schwartz_refining_2012} have pursued a categorisation of values.
Similarly, HCI literature demonstrates how ethical frameworks, like Design Justice~\cite{costanza2020design}, can guide the creation of ethical technologies.

The HRI field has investigated the ethical challenges involved in its specific context with existing frameworks~\cite{ostrowski2022ethics}.
These tools elicit the values of stakeholders -- such as roboticists, healthcare workers, and teachers -- and embed them into the design of robotic systems.
However, the emergent ethical concerns arising from the \emph{interactions} themselves and the relationships between humans and robots in the short and long term are still underexplored.

These interaction-driven values are ethical considerations that emerge during human-robot interactions, such as fairness, dependency, and autonomy.
They are context-sensitive and evolve over time, making them distinct from predefined stakeholder values.
Addressing them is crucial as they shape user experiences and the societal impact of robotic systems.
Indeed, only recently has the technology matured enough to allow overcoming challenges such as complex real-world dynamics.
Furthermore, practical tools are still needed to help researchers identify and integrate these interaction-driven values into robotic technologies.

Aiming to fill the gap in the HRI literature,
we investigate \textbf{RQ1}: \emph{what discussions on values have emerged in HRI literature to date?}
We scope our search to applications of robots for social good, in the context of healthcare, education, and private homes.
This limitation is justified by the significant societal impact and ethical complexity inherent in these contexts, where robots are increasingly integrated into personal and sensitive domains, directly influencing human wellbeing and quality of life.
In addition, while keeping into consideration the previous literature on human values, we adopt as a starting point the values arising from the interactions between humans and robots, rather than the stakeholders' values, as the latter has already been widely explored~\cite{wullenkord2020societal,ostrowski2022ethics,smakman2021moral}.

The HRI field is inherently multidisciplinary.
This can make addressing the diverse needs and perspectives from various areas of expertise challenging.
Recently, philosophers and ethics experts have become more involved in the robotic design process~\cite{axelsson2024robots}.
However, the HRI literature has yet to fully explore the potential benefits of involving these stakeholders in defining HRI values.
Therefore, our second research question is \textbf{RQ2}: \emph{what values can we define in HRI with the help of experts in the field?}

Lastly, HRI researchers often struggle to incorporate the complex and multifaceted aspects of ethics into their robot design interactions due to the complexity of the field, lack of expertise in this area~\cite{fosch2020gathering}, and absence of adequate tools that help researchers include interaction-driven ethical aspects and values in their design process~\cite{leveringhaus2018developing}.
Hence, our third research question is \textbf{RQ3}: \emph{how can we assist HRI researchers in identifying values in design scenarios?}

To address these questions, our work presents a closed-loop process consisting of three main phases as depicted in Figure~\ref{fig:steps}.
In Phase 1 -- Section~\ref{sec:phase1} -- we conducted a \textbf{scoping review} to identify the values discussed in HRI literature over the last two decades, starting from the HRI field's emergence in 2006~\cite{10.1145/1121241}, with a focus on healthcare, education, and private homes.
In Phase 2 -- Section~\ref{sec:phase2} -- we organised two \textbf{focus groups} with experts in philosophy and ethics of technology to review these values and establish a set of topics and values relevant to these HRI contexts, reported in Section~\ref{sec:values}.
Finally, in Phase 3 -- Section~\ref{sec:phase3} -- we carried out a \textbf{user study} to evaluate a tool\footnote{Available online at \url{https://hri-value-compass.github.io/}} we developed from the outcomes of the previous phases, the \textit{HRI Value Compass}, designed to help HRI researchers identify relevant values when designing robotic interfaces.
The tool references the relevant literature, offering additional insights to the researchers, and closing the process loop.

The main contributions of this work are:
\begin{enumerate}
    \item an overview of the current state of ethical discussions in the past HRI literature, that highlights the key values considered in HRI contexts;
    \item a comprehensive list of topics and values identified through a review of existing literature and consultations with experts in the field to assist the HRI research community in designing and deploying robotic systems that align with human values, social norms, and ethical principles;
    \item the development and evaluation of a design tool, the HRI Value Compass to help HRI researchers practically identify opportunities, concerns, and values when designing robotic interactions.
\end{enumerate}

\section{Related Work}

The field of HRI has been rapidly evolving, with growing attention to the social, ethical, and moral implications of robots in society.
Parallel fields have long been concerned with the ethical challenges inherent in technology design.

\subsection{Values}

\citet{schwartz_overview_2012} defines values as deeply held beliefs that are closely tied to emotions and serve to motivate actions toward desirable goals, such as independence or justice.
Unlike norms or attitudes, values transcend specific actions or situations and act as guiding standards for evaluating behaviour, policies, or events.
Although people may not always be consciously aware of their values, these values influence decisions, particularly when they come into conflict with one another.

Schwartz's theory organises values into four broad categories: openness to change, self-enhancement, conservation, and self-transcendence.
Each of these higher-order categories includes specific values, such as self-direction, power, tradition, and universalism, resulting in a total of 19 values in the revised model~\cite{schwartz_refining_2012}.
Although Schwartz's framework is valuable, it may not fully capture the context-specific values that arise in HRI, where interactions between humans and robots introduce unique ethical considerations.

\citet{graham2013moral} propose a model of moral foundations, which are psychological systems underlying human moral reasoning.
These foundations are divided into six categories: care/harm, fairness/cheating, loyalty/betrayal, authority/subversion, sanctity/degradation, and liberty/oppression. 
While this model provides insights into moral reasoning, its high-level abstractions limit its direct application in the design of technological systems, including those in HRI.
As such, identifying values specifically relevant to HRI, which may not be fully encompassed by these general models, remains crucial.

Another definition of values is the one presented by \citet{scanlon2000we} in the \emph{buck-passing} theory.
In his discourse about values, Scanlon states that ``to call something good is to claim that it has other properties [...] which provide such reasons'' for behaving in certain ways with respect to it.
The value is thus not an intrinsic property of an object or situation, but entirely depends on ``such reasons''.
We argue that since this definition imposes fewer constraints, it is better suited to model the nuanced dynamics taking place in HRI.
Consequently, the definition of value that we adopt does not follow a universal maxim and instead is more attuned to contractualist theories~\cite{scanlon1982contractualism}, which emphasise respect and consideration of individual preferences in ethical decision-making.

\subsection{Stakeholder perspectives}

Several studies have explored the perspectives of stakeholders in high-stakes environments, providing valuable insights into the ethical challenges of HRI.
\citet{cheon2016integrating} highlight the importance of including roboticists' perspectives in their design, particularly when addressing ethical concerns.
In support of this, toolkits and frameworks have been developed to assist researchers in identifying and integrating ethical values into their design processes~\cite{sadek_value-sensitive_2023}.
Design methodologies, such as Design Justice~\cite{costanza2020design}, seek to address biases in technology design and ensure that systems are developed in a manner that is fair and just.

However, while reflecting on the stakeholders' values remains fundamental, the ethical concerns and values specific to short and long-term human-robot interactions remain underexplored.
This leaves a significant gap in understanding how these systems should be ethically designed for sustainable integration into human environments.

In the field of HCI, more concrete ethical concerns have been identified.
For example, \citet{makasi_chatbot-mediated_2020} examined chatbots providing public services and identified 14 key public service values.
Although some of these values are specific to the public service context, many -- such as adaptability, fairness, accountability, and privacy -- are more broadly applicable to AI systems, including AI-powered robots.
These ethical considerations offer important insights for the design of systems that interact directly with humans, such as social robots.

In the context of robotics, \citet{fosch2020gathering} analysed discussions with experts, revealing key ethical, legal, and social challenges in the deployment of social robots. The authors identified five primary areas of concern: security and privacy, legal uncertainty, autonomy and agency, employment, and the replacement of human interaction. These concerns are particularly pertinent to robots designed for social good, such as those used in healthcare, education, and domestic settings. In our work, we aim to build on these insights by identifying concerns and values specific to HRI, providing a comprehensive list of values that can guide the design of robotic systems that align with human values and societal norms.

\citet{ostrowski2022ethics} conducted a comprehensive review of HRI literature over the past two decades, examining various interpretations of ethics within the field.
One particularly relevant aspect of their work is the exploration of ethics in human-robot relationships.
In this context, they successfully identified several key ethical themes, which we aim to further validate and expand upon in our own research.

\section{Phase 1: Scoping Review}
\label{sec:phase1}

The main objective of this phase is to understand the key discussions revolving around values that have emerged in the HRI field in the last two decades (\textbf{RQ1}). 
To achieve this, we conducted a scoping review focused on identifying and extracting the concerns and values from HRI studies published since 2006, when the ACM/IEEE International Conference on Human-Robot Interaction was established~\cite{10.1145/1121241}.

\subsection{PRISMA process}

\begin{figure}[t]
    \centering
    \includegraphics[width=0.8\columnwidth]{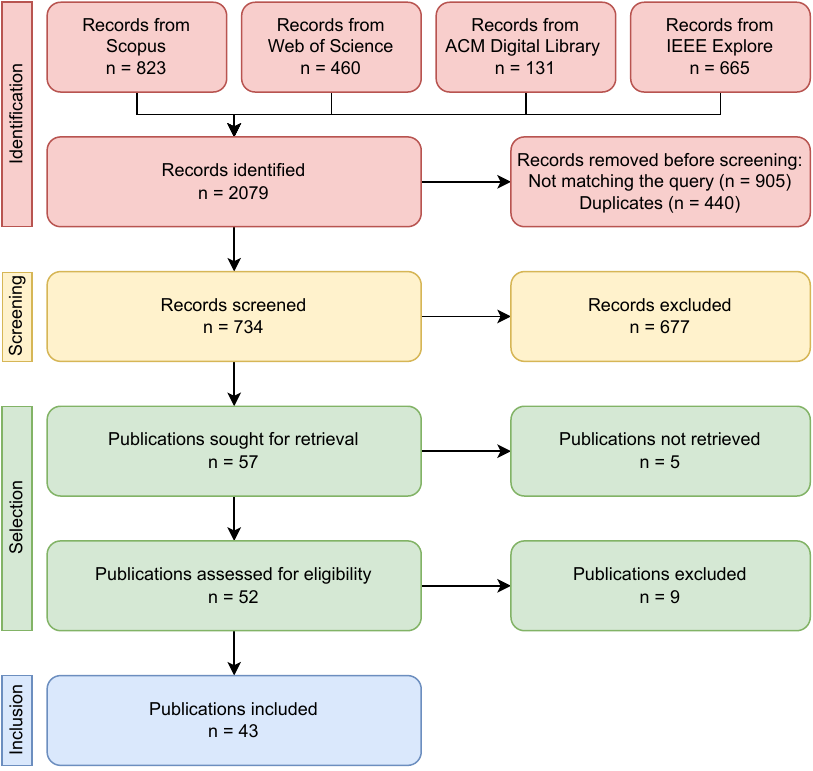}
    \caption{PRISMA framework flow diagram for the scoping review.}
    \label{fig:prisma}
\end{figure}

We followed the PRISMA framework~\cite{page2021prisma}, the state-of-the-art workflow for systematic literature reviews, adapted for a scoping review~\cite{tricco2018prisma}.
The process includes the following steps, as shown in Figure~\ref{fig:prisma}: (i) \emph{identification}, searching specific databases within a defined time range, (ii) \emph{screening}, filtering out records whose titles and abstracts do not meet the eligibility criteria, (iii) \emph{selection}, reading the full texts of the publications and excluding those that do not meet the eligibility criteria, and (iv) \emph{inclusion}, selecting the final list of publications and extracting their data.

\subsubsection{Search query}

Based on the research question \textbf{RQ1} and the scope of this work, we constructed a search query as in~\cite{catania2023conversational, spitale2023systematic}.
The query is satisfied for a record if the combination of the record's title, abstract and keywords satisfies the following: contains the term \emph{social}; contains the term \emph{robot} or its variations; contains any of the terms \emph{value}, \emph{ethic}, \emph{moral}, \emph{deont-} (for deontology); and contains any of the terms \emph{school}, \emph{teach}, \emph{education}, \emph{assist}, \emph{healthcare}, \emph{patient}, \emph{home}, \emph{house}, \emph{domestic}.

We collected the records that satisfied the query on March 27th 2024, across Scopus, Web of Science, ACM Digital Library, and IEEE Explore databases.
These databases were selected as they include works from major publishers and cover most journals and venues that publish works related to HRI.
When allowed by the querying tool, we filtered for research articles in journals or proceedings, published between January 1st 2006 and December 31st 2023, and written in English.
Although we recognise that this excludes works published in 2024, our decision to focus on publications up to 2023 ensures that the analysis is based on a complete and stable body of literature. Furthermore, works from the most recent year would have limited influence on broader trends.
Where a database querying tool did not have the required filtering granularity level, we used a less strict query.
After collecting the records, we filtered them again with a custom script implementing the search query exactly, ensuring uniform results.
The records excluded during this step are reported in Figure~\ref{fig:prisma} as ``Not matching the query''.

\subsubsection{Exclusion criteria}

During the selection process, we followed specific inclusion criteria and the records were excluded if: 
\begin{itemize}
    \item no physical robot was present (e.g., virtual agents are excluded);
    \item the scope of the contribution was not about robots for social good (e.g., service robots, physical assistive technologies and transport robots are excluded);
    \item the contribution was a literature review or a survey;
    \item the topic was not about education, care, or domestic environment;
    \item the publication cannot be retrieved.
\end{itemize}

\subsubsection{Selection process}

The results of the research query from the four databases were merged obtaining 2079 records; after filtering and duplicates removal, the remaining records were 734, as reported in Figure~\ref{fig:prisma}.

During the screening step, the records were randomly divided into two groups and assigned to two reviewers (male, 27 years old, female, 31 years old) who filtered them according to the exclusion criteria.
In this step, the reviewers considered only the title and abstract associated with each record.
If a reviewer was unsure about a paper, they flagged it for further discussion with all reviewers.
Before starting the screening process, we randomly sampled 10 papers and the two reviewers screened them, to ensure consensus on inclusion and exclusion criteria.
The resulting set contained 57 records to be retrieved, of which 5 were unavailable and thus have been excluded.

Subsequently, the retrieved publications were again split randomly into two sets and assigned to the same reviewers, who read the full texts for the final selection and data extraction.
Again, if a reviewer was uncertain about including a paper, this was discussed with the other reviewer for a final decision.
The final selection comprises 43 publications and a list of the publications is available in Tables~\ref{tab:phase1-output} and~\ref{tab:phase1-output-2} and also in a publicly accessible table stored in our web tool\footnote{\url{https://hri-value-compass.github.io/}}.

\subsubsection{Data extraction and analysis}

\begin{table}[t]
    \caption{Variables extracted from the publications included in the study.}
    \label{tab:variables}
    \begin{adjustbox}{width=\textwidth,center=\textwidth}
    \footnotesize
    \begin{tabular}{lll}
\hline
\textbf{Variable}            & \textbf{Type} & \textbf{Example}                                   \\
\hline
Publication's year           & Numerical     & 2010, ...                                          \\
Application topic            & Categorical   & healthcare, education, ...                         \\
Authors' affiliation country & Categorical   & USA, Germany, Europe, ...                          \\ 
Empirical study              & Boolean       & 0, 1                                               \\ 
Population's country         & Categorical   & United Kingdom, ...                                \\
Population's age             & Categorical   & Children, Elderly, ...                             \\ 
Transcripts                  & Qualitative   & ``the issues of infantilisation of elderly [...]'' \\
\hline
\end{tabular}
    \end{adjustbox}
\end{table}

We defined a set of variables to be extracted from each of the included publications.
These variables were selected based on the goal of the scoping review, i.e., obtaining a snapshot of the values that emerged in the HRI field over the last two decades.
Since these values are subjective and vary across cultures, we included information on the author's country of affiliation and on whether an empirical study was performed.
Where applicable, we recorded the nationality and age of the participants involved.

In our procedure, we extracted data solely from the main content of each publication, without including the supplementary material.
This ensured a consistent extraction procedure across all papers. 
The variables collected, reported in Table~\ref{tab:variables}, are of different types: numerical (e.g., year of the publication), categorical (e.g., countries), or qualitative (e.g., transcripts).

For extracting the qualitative variables, we adopted the thematic analysis approach~\cite{braun2012thematic}, which consists of the following six steps.
During the data extraction phase described above, for each paper assigned to them, each of the reviewers (Step 1) familiarised themselves with the data by copying verbatim parts of the papers (transcripts) that were relevant to the scope.
Subsequently, they (Step 2) generated the initial codes and (Step 3) searched for themes by collating codes.
Then, together, they (Step 4) reviewed the themes, comparing their results and discussing the differences, (Step 5) named the themes into meaningful values according to the literature, and (Step 6) produced a report.

\subsection{Results: value-related HRI papers}

\begin{figure}[t]
    \centering
    \includegraphics[width=0.43\linewidth]{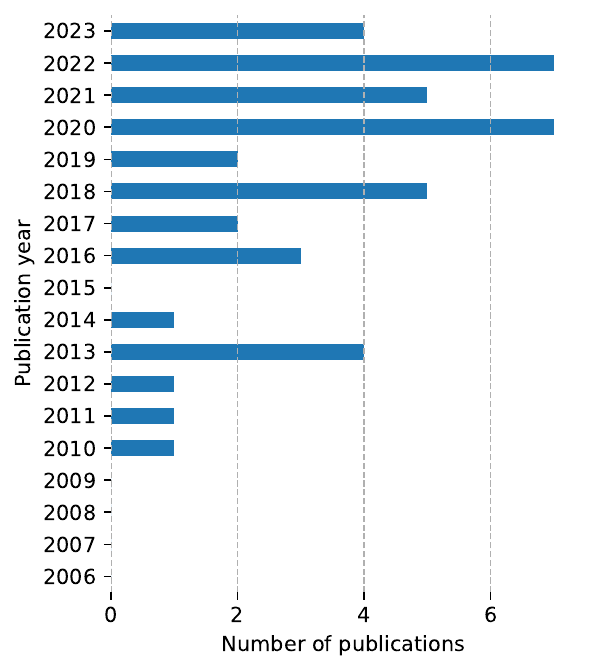}
    \includegraphics[width=0.50\linewidth]{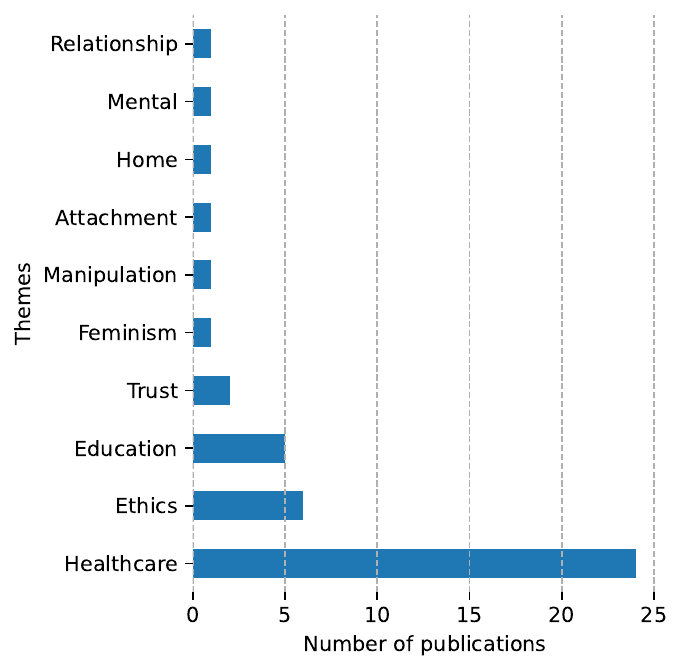}
    \caption{On the left: distribution of the publication years of the selected papers; on the right: occurrences of each application topic in the selected publications.}
    \label{fig:publication-years}
\end{figure}

\begin{table*}[ht!]
    \caption{Initial values identified in the scoping review. Refer to Table~\ref{tab:final_values} for the final version.}
    \label{tab:initial_values}
    \begin{adjustbox}{width=1.2\textwidth,center=\textwidth}
    \footnotesize
    \begin{tabular}{lllp{0.5\linewidth}}
\hline
\textbf{Value} & \textbf{Topic - Positive} & \textbf{Topic - Negative} & \textbf{Description}                                                                            \\
\hline
autonomy      & independence     &                   & The user becomes independent from other humans; the user is independent in using the robot.                       \\
autonomy      &                  & attachment        & The user can develop an emotional dependency on the robot.                                                        \\
autonomy      &                  & dependency        & The user can develop a physical dependency on the robot.                                                          \\
connectedness & companionship    &                   & The robot can provide companionship.                                                                              \\
connectedness & social enabler   &                   & The robot can facilitate connections between humans.                                                              \\
connectedness &                  & social substitute & The robot can reduce social interactions.                                                                         \\
empowerment   & access           &                   & The users or others can have access to the data collected by the robot.                                           \\
empowerment   & control          &                   & The human remains in the loop; the robot does not constrain or force the user.                                    \\
empowerment   &                  & manipulation      & The robot can influence the user; the robot can lie.                                                              \\
equality      & fairness         &                   & The robot must treat everyone equally.                                                                            \\
equality      &                  & exclusion         & Some users do not feel part of the intended audience; the robot is not personalised; the robot is not affordable. \\
honour        &                  & agency            & The robot should not trick the users into anthropomorphising it.                                                  \\
honour        &                  & degradation       & The users might perceive interacting with a robot as humiliating.                                                 \\
kindness      & mentor           &                   & The robot can set a good example and foster virtuous interactions.                                                \\
kindness      &                  & corruptor         & The users can mistreat the robot, fostering negative interactions.                                                \\
privacy       &                  & consent           & The robot invades bystanders' privacy.                                                                            \\
privacy       &                  & intimacy          & The robot's constant presence invades private spaces.                                                             \\
privacy       &                  & profiling         & The robot collects data continuously or aggressively; how the data is protected.                                  \\
truth         & accountability   &                   & Who is responsible for the robot's actions.                                                                       \\
truth         & transparency     &                   & The robot should make its artificial nature clear, and inform about monitoring and functionalities.               \\
welfare       & education        &                   & The robot can teach topics, thought processes, and routines.                                                      \\
welfare       & nonjudgmental    &                   & The user can be sincere with the robot, as it has no sentiments.                                                  \\
welfare       & safety           &                   & The robot must not harm the users.                                                                                \\
welfare       & wellbeing        &                   & The robot allows disclosure and can give advice.                                                                  \\
\hline
\end{tabular}
    \end{adjustbox}
\end{table*}

\begin{table*}[ht!]
    \caption{List of papers included in our survey and corresponding values and topics extracted. Note that this table has been updated with the final values defined at the end of the design process to avoid any confusion. [Continued in Table~\ref{tab:phase1-output-2}].}
    \label{tab:phase1-output}
    \begin{adjustbox}{width=1.4\textwidth,center=\textwidth}
    \tiny
    \centering
    \begin{tabular}{p{0.2\linewidth}lllp{0.29\linewidth}p{0.2\linewidth}}
    \toprule
    \textbf{Authors} & \textbf{Year} & \textbf{Application} & \textbf{Study} & \textbf{Topics} & \textbf{Values} \\
    \midrule
    \citet{sharkey2010ethical} & 2010 & Healthcare & No & Control, Intimacy, Social enabler, Social substitute, Tricking & Agency, Connectedness, Dignity, Privacy \\
    \citet{calo2011ethical} & 2011 & Healthcare & No & Attachment, Expectations, Manipulation, Meta, Social enabler, Social substitute, Tricking & Autonomy, Connectedness, Dignity \\
    \citet{sharkey2012granny} & 2012 & Healthcare & No & Independence, Intimacy, Safety, Social substitute, Tricking & Agency, Connectedness, Dignity, Privacy, Welfare \\
    \citet{amirabdollahian2013accompany} & 2013 & Healthcare & Yes & Independence, Safety, Social enabler & Agency, Connectedness, Welfare \\
    \citet{amirabdollahian2013assistive} & 2013 & Healthcare & No & Meta, Profiling, Safety & Privacy, Welfare \\
    \citet{misselhorn2013ethical} & 2013 & Healthcare & No & Degradation, Social substitute, Tricking, Wellbeing & Connectedness, Dignity, Welfare \\
    \citet{van2013method} & 2013 & Ethics & No & Accountability, Control, Meta & Agency, Equity \\
    \citet{fridin2014kindergarten} & 2014 & Education & Yes & Attachment, Profiling, Safety & Autonomy, Privacy, Welfare \\
    \citet{huber2016ethical} & 2016 & Attachment & No & Dependence, Identity disclosure, Profiling & Agency, Dignity, Privacy \\
    \citet{ienca2016social} & 2016 & Healthcare & No & Meta, Safety, Transparency & Privacy, Welfare \\
    \citet{sharkey2016should} & 2016 & Healthcare & No & Attachment, Identity disclosure, Profiling, Social substitute, Transparency & Autonomy, Connectedness, Dignity, Privacy \\
    \citet{aldinhas2017robots} & 2017 & Healthcare & No & Intimacy, Profiling, Social substitute & Connectedness, Privacy \\
    \citet{draper2017ethical} & 2017 & Healthcare & Yes & Access, Accountability, Control, Dependence, Expectations, Manipulation, Meta, Social substitute, Wellbeing & Agency, Autonomy, Connectedness, Equity, Privacy, Welfare \\
    \citet{battistuzzi2018embedding} & 2018 & Healthcare & No & Companionship, Independence, Intimacy, Tricking & Agency, Connectedness, Dignity, Privacy \\
    \citet{cresswell2018health} & 2018 & Healthcare & Yes & Control, Safety, Tricking & Agency, Dignity, Welfare \\
    \citet{garcia2018inclusion} & 2018 & Healthcare & Yes & Companionship & Connectedness \\
    \citet{poulsen2018care} & 2018 & Trust & Yes & Identity disclosure, Transparency & Dignity, Privacy \\
    \citet{vcaic2018service} & 2018 & Healthcare & Yes & Assistance, Companionship, Dependence, Profiling, Safety, Social enabler, Social substitute & Agency, Connectedness, Privacy, Welfare \\
    \citet{fiske2019your} & 2019 & Mental & No & Accountability, Attachment, Expectations, Manipulation, Meta, Profiling, Social substitute, Tricking & Autonomy, Connectedness, Dignity, Equity, Privacy \\
    \citet{li2019perceptions} & 2019 & Home & Yes & Dilemma, Meta & N/A \\
    \citet{malle2020general} & 2020 & Ethics & No & Meta & N/A \\
    \citet{poulsen2020cybersecurity} & 2020 & Healthcare & Yes & Profiling, Safety & Privacy, Welfare \\
    \citet{smakman2020acceptable} & 2020 & Education & Yes & Accountability, Assistance, Attachment, Control, Education, Manipulation, Mentor, Nonjudgmental, Profiling, Social substitute, Tricking & Agency, Autonomy, Connectedness, Dignity, Equity, Privacy, Virtue, Welfare \\
    \citet{tuuli2020use} & 2020 & Healthcare & Yes & Expectations, Meta & N/A \\
    \citet{van2020teachers} & 2020 & Education & Yes & Accountability, Consent, Manipulation, Social substitute, Transparency, Wellbeing & Autonomy, Connectedness, Equity, Privacy, Welfare \\
    \citet{vandemeulebroucke2020ethics} & 2020 & Ethics & Yes & Accountability, Bias, Control, Degradation, Dependence, Exclusion, Independence, Intimacy, Meta, Profiling, Safety, Social enabler, Social substitute, Tricking, Wellbeing & Agency, Connectedness, Dignity, Equity, Privacy, Welfare \\
    \citet{vanderelst2020can} & 2020 & Healthcare & Yes & Accountability, Control, Expectations, Profiling & Agency, Equity, Privacy \\
    \citet{cappuccio2021can} & 2021 & Ethics & No & Bias, Corruptor, Dependence, Exclusion, Expectations, Fairness, Mentor & Agency, Equity, Virtue \\
    \bottomrule
\end{tabular}

    \end{adjustbox}
\end{table*}

\begin{table*}[ht!]
    \caption{[Continued from Table~\ref{tab:phase1-output}] List of papers included in our survey and corresponding values and topics extracted. Note that this table has been updated with the final values defined at the end of the design process to avoid any confusion.}
    \label{tab:phase1-output-2}
    \begin{adjustbox}{width=1.4\textwidth,center=\textwidth}
    \tiny
    \centering
    \begin{tabular}{p{0.2\linewidth}lllp{0.29\linewidth}p{0.2\linewidth}}
    \toprule
    \textbf{Authors} &  \textbf{Year} & \textbf{Application} & \textbf{Study} & \textbf{Topics} & \textbf{Values} \\ 
    \midrule
    \citet{malinverni2021educational} & 2021 & Education & Yes & Wellbeing & Welfare \\
    \citet{prescott2021friends} & 2021 & Relationship & No & Companionship, Degradation, Expectations, Safety, Social enabler, Social substitute, Wellbeing & Connectedness, Dignity, Welfare \\
    \citet{smakman2021moral} & 2021 & Education & Yes & Access, Accountability, Attachment, Consent, Education, Exclusion, Expectations, Fairness, Independence, Manipulation, Mentor, Profiling, Social substitute, Transparency, Wellbeing & Agency, Autonomy, Connectedness, Equity, Privacy, Virtue, Welfare \\
    \citet{wagman2021beyond} & 2021 & Feminism & No & Bias, Exclusion, Expectations, Fairness, Manipulation & Autonomy, Equity \\
    \citet{baisch2023elders} & 2022 & Healthcare & Yes & Attachment, Bias, Expectations & Autonomy, Equity \\
    \citet{carros2022ethical} & 2022 & Ethics & Yes & Dilemma, Expectations, Meta & N/A \\
    \citet{felber2022concept} & 2022 & Healthcare & No & Accountability, Degradation, Exclusion, Profiling, Social substitute & Connectedness, Dignity, Equity, Privacy \\
    \citet{hungtechnological} & 2022 & Healthcare & Yes & Consent, Social enabler & Connectedness, Privacy \\
    \citet{ostrowski2022mixed} & 2022 & Healthcare & Yes & Attachment, Independence, Profiling, Transparency & Agency, Autonomy, Privacy \\
    \citet{street2022older} & 2022 & Healthcare & Yes & Control, Corruptor, Dependence, Identity disclosure, Independence, Profiling, Social substitute & Agency, Connectedness, Dignity, Privacy, Virtue \\
    \citet{wheeler2023confucianism} & 2022 & Healthcare & No & Exclusion, Transparency, Wellbeing & Equity, Privacy, Welfare \\
    \citet{jung2023enjoy} & 2023 & Manipulation & Yes & Attachment, Companionship, Education, Exclusion, Manipulation, Transparency & Autonomy, Connectedness, Equity, Privacy, Welfare \\
    \citet{rosero2023using} & 2023 & Trust & Yes & Dilemma, Identity disclosure & Dignity \\
    \citet{soares2023ethics} & 2023 & Healthcare & Yes & Accountability, Expectations, Independence & Agency, Equity \\
    \citet{yeung2023living} & 2023 & Ethics & Yes & Mentor, Nonjudgmental, Social enabler, Social substitute, Wellbeing & Connectedness, Virtue, Welfare\\
    \bottomrule
\end{tabular}

    \end{adjustbox}
\end{table*}

The analysis of the \textbf{publication's year} of the 43 selected papers showed an increase in the number of contributions to the topic over time, as evident in Figure~\ref{fig:publication-years}.
This reflects the community's crescent interest in exploring and becoming aware of the values that drive the interactions between humans and robots.

\subsubsection{Application topics}

The paper's \textbf{application topics} were extracted with the thematic analysis approach~\cite{braun2012thematic} as in previous works~\cite{spitale2023systematic, catania2023conversational}, resulting in 10 areas, reported in Figure~\ref{fig:publication-years}.
24 papers reported works on healthcare (e.g., elderly healthcare~\cite{amirabdollahian2013assistive}), 6 on ethics (e.g.,~\cite{cappuccio2021can}), 5 on education (e.g.,~\cite{van2020teachers}), 2 on trust (e.g.,~\cite{rosero2023using}), 1 on feminism, 1 on manipulation, 1 on attachment, 1 on home, 1 on mental health, and 1 on human-robot relationships. 

\subsubsection{Cultural background}

The \textbf{country of affiliation of the authors} of the publications is from North America and Western Europe.
While the country of affiliation of the authors is not necessarily indicative of their cultural background, this information provides an idea of the point of view of the contributions included in the review, which is Western-centric.
Of the 43 publications considered, 25 featured an \textbf{empirical study}.
In these cases, knowing the \textbf{population's country} of origin, or the place where the study was executed, is relevant. This geographical distribution mirrors the authors' case, although it includes data from Brazil and China.
Regarding the \textbf{population's age}, the studies encompass young, adult and elderly people interacting with social robots, and in three cases they include other stakeholders, e.g., parents, teachers, and caretakers.

\subsubsection{Transcripts}
\label{sec:thematic-analysis}

The two reviewers extracted a total of 235 \textbf{transcripts} from the 43 surveyed papers, each labelled with a code (as for the thematic analysis approach), which we will refer to as \emph{topics}.
After an iterative process of distillation of the topics into themes and a renaming of these based on the existing literature, we extracted 24 initial topics grouped under 9 initial themes, which we will refer to as \emph{values}.
The topics and values have been further refined during the second step of this research, detailed in Section~\ref{sec:phase2}.
For a complete description of the final topics and values that emerged from the HRI literature, please refer to Section~\ref{sec:values}.

In addition, the topics have a positive or negative valence.
For instance, the topic \emph{companionship}, indicating that a robot can provide comfort and entertainment for lonely users, has a positive valence.
Indeed, all the transcripts from literature grouped under this topic present this with positive connotations.
When a similar concept is presented with negative connotations, a different topic is used.
For example, the fact that a companion robot can reduce the user's social interactions with other humans was grouped under a topic with negative valence, namely, \emph{social substitute}.

We identified four topics outside our work's scope.
We labelled as \emph{dilemma} the excerpts reflecting those challenging moral dilemmas such as the trolley cart problem.
With \emph{meta} we labelled those contributions that offer insights, such as terms' definitions, but do not focus on the interactions; these were instrumental in choosing the terms we adopted.
Finally, under \emph{expectations} and \emph{assistance} we grouped the considerations about what the public expects from robots, and how the robots can be used as personal assistants.

We report the initial topics and values in Table~\ref{tab:initial_values} while Tables~\ref{tab:phase1-output} and~\ref{tab:phase1-output-2} collect the papers included in the scoping review and the corresponding values and topic defined at the end of the design process are presented in detail in Section~\ref{sec:values}.

\subsection{Preliminary discussion and limitations}
Results from Phase 1 contribute to the field of HRI by offering an overview of the current ethical discussions around values in the context of healthcare, education, and domestic HRI.
Within application contexts, healthcare emerged as the most represented domain, suggesting that ethical considerations are especially relevant in sensitive contexts where robots directly impact human health and wellbeing. Our results show that other topics like feminism and manipulation were largely underexplored. However, this could have been affected by the choice of the search query that only included domains of healthcare, education, and domestic. Future research could investigate these areas to provide a more holistic view of ethical considerations in HRI.
The Western-centric perspective of the surveyed publications points to a significant gap in the cultural representation of values in HRI. Future work should explore more cultural diversity in different socio-cultural contexts.
While we chose the interaction as a starting point for the value identification, the results are influenced by other elements, such as the stakeholders' values and socio-cultural aspects, with effects on the robot's role beyond the interaction itself.

\section{Phase 2: Focus Groups}
\label{sec:phase2}

Building on the list of values that emerged from the literature review, Phase 2 aims to define a comprehensive list of topics and values specific to human-robot interaction (\textbf{RQ2}). To achieve this, we conducted two focus groups with six experts in technology ethics and philosophy, consisting of a discussion driven by three activities, to develop a list of values in the context of HRI.
This approach aligns with previous research, such as \citet{pommeranz2012social}, which successfully used expert focus groups to explore complex themes in technology and negotiation.

\subsection{Participants}

We invited a total of fourteen (7 females, and 7 males) experts in technology ethics and philosophy from various parts of Europe, by sending email invitations for voluntary participation in the focus groups. Four male  (P1--P4) and two female (P5--P6) experts agreed to participate.
Among them, are a professor in Robotics, AI and Ethics of Technology, a lecturer in Philosophy of Technology, three postdoctoral researchers in Ethics of Information Technology and Moral Reasoning in AI Systems, and a professor in Ethics of Technology and Philosophy. Due to participant scheduling constraints, we organised two focus groups. The first group included participants P1, P2, and P4, while the second group comprised P5 and P6.
The participants were introduced to the study objectives and they signed an informed consent form, following Ghent University Ethics Committee’s guidelines.

\subsection{Focus group activities}

We conducted the focus groups, which lasted approximately two hours, using an online video conference tool; the sessions were recorded for later analysis.
To facilitate collaboration, we used Miro\footnote{\url{https://miro.com/}}, an online platform that allows users to interact with interactive boards that can be prepared in advance.
The focus groups consisted of three structured activities: (1) HRI scenario analyses (individual task), (2) values review (group task), and (3) contextual information and relationship with values (group task).

These activities, designed to leverage the dynamics of group discussion while ensuring individual reflection, are inspired by previous research in related fields.
The use of scenarios and visual aids in focus groups is supported by HCI research, which demonstrates that such tools stimulate richer discussions~\cite{goodman_Gathering_Requirements}.
Our methodology applies it to the HRI field as in \citet{spitale2024appropriateness}, who used a scenario-based evaluation to assess whether the language generated by large language models was appropriate for a robotic coach promoting mental wellbeing.

Each session was guided by two researchers: one introduced the activities and acted as a moderator to facilitate discussion and stimulate conversation, without interrupting the discussion flow of the participants and avoiding imposing external views, while the other observed and took detailed notes.
This setup was chosen to minimise researcher influence on participant interactions.

\subsubsection{Activity 1: scenarios analysis}
\label{sec:step2act1}

The first activity involved analysing four distinct HRI fictional value scenarios provided to the participants. A value scenario is an extension of scenario-based design that helps envision the systemic effects of new technologies~\cite{nathan2007value}.
We developed our HRI value scenarios through a structured process.
First, we identified the main topics for our value scenarios based on the results of the previous phase and previous research~\cite{lambert2020systematic}.
Specifically, we considered healthcare (focused both on elderly care~\cite{broekens2009assistive} and autism therapy~\cite{diehl2012clinical}), education (focused on children's learning~\cite{mubin2013review}), and home assistance~\cite{sung2008housewives} as our application contexts (matching the keywords of the search query in Phase 1).
Second, we created four value scenarios, one for each identified HRI domain, using ChatGPT\footnote{\url{https://chatgpt.com/}}, as demonstrated by \citet{jung2023toward}. We followed the same approach to draft prompts for generating these scenarios. The specific prompts used to generate the scenarios are provided in Appendix~\ref{app:prompt}.
Third, we refined the generated value scenarios by removing any references to potential moral values to avoid biasing the scenarios.
The scenarios generated are reported in Appendix~\ref{app:scenarios}.

We provided participants with these value scenarios one by one and asked them to individually identify the moral values present in each scenario.
The participants provided written responses on virtual sticky notes in Miro. This individual reflection phase was designed to ensure all participants, including less confident members, had the opportunity to articulate their thoughts before the group discussion. Following this, the moderator initiated a group discussion, encouraging participants to share and debate their perspectives.
This activity lasted approximately 40 minutes, 10 minutes for each scenario.

\subsubsection{Activity 2: values review}

The second activity involved presenting participants with the results from Phase 1, which included a list of topics and values identified and extracted from the literature. To make these HRI moral values easier to visualise, we organised them in both a table (see Table~\ref{tab:initial_values}) and a wheel (as shown in Figure~\ref{fig:values-wheel}).
We first asked participants to take their time reviewing the values, then encouraged them to collaboratively discuss and reflect on them.
Their opinions were transcribed in real time in Miro by one of the researchers to support the discussion.
This activity lasted approximately 20 minutes.

\subsubsection{Activity 3: contextual information and relationship with values}
\label{sec:act3}

\begin{figure*}[t]
    \centering
    \includegraphics[width=\textwidth]{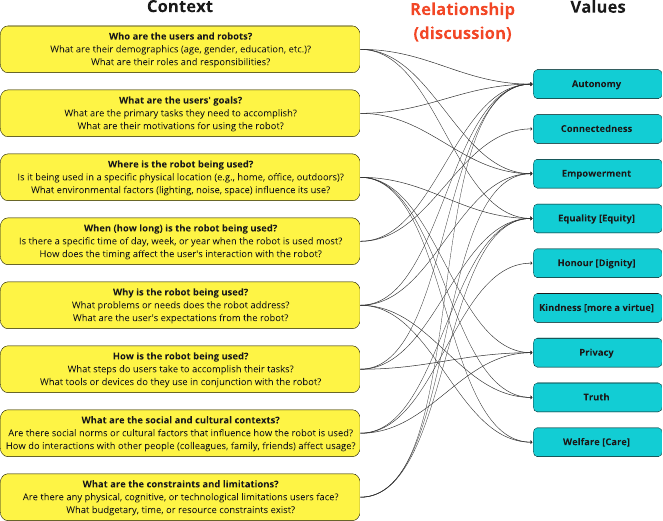}
    \caption{Miro board screenshot of activity 3 in which participants were asked to discuss the relationships between contextual information and values at play.}
    \label{fig:activity3}
\end{figure*}

The final activity involved discussing the relationship between the identified values and the contextual information. Our literature review revealed that most values are highly context-dependent in line with previous literature~\cite{de2016ethical}. Therefore, we decided to prioritise context as the main lens for framing these values more effectively in any HRI scenario. To this end, we have designed an activity where we asked participants to discuss the relationship between the values they have reviewed and the contextual information of HRI scenarios taking inspiration from theories in Contextual Design~\cite{holtzblatt1997contextual}. We specifically prompted them using the following contextual questions (as depicted in Figure~\ref{fig:activity3}): (1) \textit{Who} are the users and robots? What are their demographics (age, gender, education, etc.)? What are their roles and responsibilities?; (2) \textit{What} are the users' goals? What are the primary tasks they need to accomplish? What are their motivations for using the robot?; (3) \textit{Where} is the robot being used? Is it being used in a specific physical location (e.g., home, office, outdoors)? What environmental factors (lighting, noise, space) influence its use?; (4) \textit{When} (how long) is the robot being used? Is there a specific time of day, week, or year when the robot is used most? How does the timing affect the user's interaction with the robot?; (5) \textit{Why} is the robot being used? What problems or needs does the robot address? What are the user's expectations from the robot?; (6) \textit{How} is the robot being used? What steps do users take to accomplish their tasks? What tools or devices do they use in conjunction with the robot?; (7) \textit{What} are the social and cultural \textit{contexts}? Are there social norms or cultural factors that influence how the robot is used? How do interactions with other people (colleagues, family, friends) affect usage?; (8) \textit{What} are the \textit{constraints} and limitations? Are there any physical, cognitive, or technological limitations users face? What budgetary, time, or resource constraints exist?
This activity lasted around one hour.

\subsubsection{Data analysis}
We first automatically transcribed the two focus group recordings and then manually reviewed them for accuracy. We used the thematic analysis method~\cite{braun2012thematic}, with the values identified during the scoping review serving as main themes and the topics as codes. 

\subsection{Results: topics and values in HRI}
\label{sec:values}

\begin{table}[ht!]
    \caption{Final topics and values identified in the scoping review, updated in accordance with the results of the focus groups.}
    \label{tab:final_values}
    \begin{adjustbox}{width=1.2\textwidth,center=\textwidth}
    \footnotesize
    \begin{tabular}{lllp{0.5\linewidth}}
\hline
\textbf{Value} & \textbf{Topic - Positive} & \textbf{Topic - Negative} & \textbf{Description}                                                                               \\
\hline
agency        & independence        &                   & The robot reduces the user's physical dependence on other humans; the user is independent in using the robot.     \\
agency        &                     & dependence        & The user can develop a physical dependence on the robot.                                                          \\
agency        & control             &                   & The robot does not physically constrain or force the user.                                                        \\
connectedness & companionship       &                   & The robot can provide companionship.                                                                              \\
connectedness & social enabler      &                   & The robot can facilitate connections between humans.                                                              \\
connectedness &                     & social substitute & The robot can reduce social interactions.                                                                         \\
privacy       & access              &                   & The users, or people selected by them, can have access to the data collected by the robot.                        \\
privacy       &                     & consent           & The robot invades bystanders' privacy.                                                                            \\
privacy       &                     & intimacy          & The robot's constant presence invades private spaces, causing discomfort and a panopticon effect.                 \\
privacy       &                     & profiling         & The robot collects data continuously or aggressively; data safety and protection.                                 \\
privacy       & transparency        &                   & The robot's functionalities and monitoring capabilities are clearly stated.                                       \\
autonomy      &                     & attachment        & The user can develop an emotional dependency on the robot, reducing the autonomy.                                 \\
autonomy      &                     & manipulation      & The robot can influence the user with persuasion and nudging.                                                     \\
equity        & fairness            &                   & The robot treats everyone fairly according to their needs.                                                        \\
equity        &                     & bias              & Some users do not feel part of the intended audience (algorithmic bias).                                          \\
equity        &                     & exclusion         & The robot is not personalised to the user demographic: preferences, habits, economic background...                \\
equity        & accountability      &                   & The person responsible for the robot's actions is fairly accountable.                                             \\
dignity       &                     & tricking          & The robot should not lie or trick the users into anthropomorphising its behaviour.                                \\
dignity       &                     & degradation       & The users might perceive interacting with a robot instead of a human as humiliating.                              \\
dignity       & identity disclosure &                   & The robot's artificial nature is clearly stated.                                                                  \\
virtue        & mentor              &                   & The robot can set a good example and foster virtuous interactions.                                                \\
virtue        &                     & corruptor         & The users can exercise unkind behaviour towards the robot, with negative effects on subsequent human interactions.\\
welfare       & education           &                   & The robot can teach topics, thought processes, and routines.                                                      \\
welfare       & nonjudgmental       &                   & The user can be sincere with the robot, as it has no sentiments.                                                  \\
welfare       & safety              &                   & The robot is safe and it does not harm the users.                                                                 \\
welfare       & wellbeing           &                   & The robot allows disclosure and can give advice.                                                                  \\     
\hline
\end{tabular}
    \end{adjustbox}
\end{table}

\begin{figure}[htb!]
    \centering
    \includegraphics[width=0.7\linewidth]{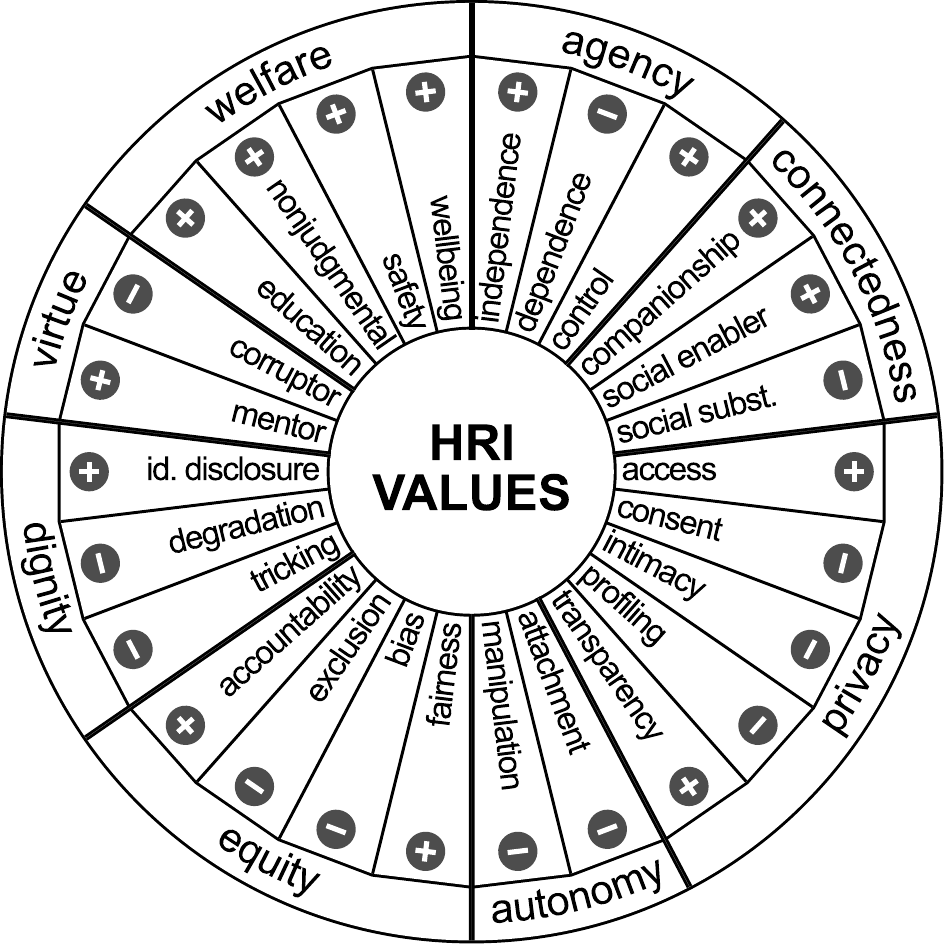}
    \caption{Final topics and values, with positive and negative valences.}
    \label{fig:values-wheel}
\end{figure}

In Activity 1, the participants were asked to identify the values in the fictional HRI scenarios presented to them, without any influence from our preliminary scoping review findings. 
Activity 2 revealed the gaps in the values identified by the scoping review, while Activity 3 highlighted the link between contextual information and moral values. The participants noted that each piece of contextual information could be associated with multiple moral values, though they focused on the most relevant connections during discussions. At the end of the focus group, we revised the initial value definitions from the scoping review, and based on the outputs of the focus groups, we defined 26 topics grouped under 8 values, as shown in Table~\ref{tab:final_values}. Each topic has a positive or negative valence and is visualised in Figure~\ref{fig:values-wheel}.
For clarity, we combined the results from all activities and we report the final list in the following sections.
Appendix \ref{app:phase2-results} provides a detailed overview of the thematic analysis, including citations, quotes, codes, and themes from participants organised by activity. Our analysis reached data saturation \cite{hennink2022sample}, allowing us to confidently identify and understand the final list of values, which are described in the following paragraphs.
Note that this is a simplified and reductionist framework designed to assist researchers in identifying and addressing values within the HRI scenarios they are developing and not a comprehensive tool that captures the full complexity of ethical concerns and values, cultural contexts, and stakeholders' perspectives in the HRI realm.

\subsubsection{Value: Agency}

This value encompasses three topics -- independence, dependence, and control -- all related to the user's physical freedom. 
This term, prior to the focus groups, referred to one of the topics, but it was promoted to a value following the suggestions from the focus groups participants, uniforming our definition of agency with \citet{prunkl2022human}: ``A person is able to act on the beliefs and values they hold. This implies that they have meaningful options available to them, allowing them to make choices that are of practical import to their life''.
As it emerged during the focus groups, agency differs from autonomy as the latter is used in this work to encompass topics which do not possess physical connotations. The related topics are defined as follows.

\paragraph{Topic: Independence}

The use of robots in daily life can significantly enhance a user's independence, by reducing their physical dependence on other people, contributing to a sense of personal agency.
This is especially evident in the case of elderly individuals 
or people with physical disabilities 
who require constant assistance. 
We included in this topic the topics about ease of use of the robot appliance, which should not require the user to rely on someone else to use the robot, as emerged during the scoping review~\cite{street2022older, vandemeulebroucke2020ethics}.

\paragraph{Topic: Dependence}

The previous topic can also be viewed with negative connotations~\cite{cappuccio2021can, vcaic2018service}.
Indeed, when the subject's independence relies on robots instead of other humans, the subject becomes dependent on the robots themselves. For example, the focus groups' participants mentioned that this could lead to a deterioration of various skills, such as cooking or reasoning. 
One of the participants particularly questioned if deskilling might be offset by the development of new, possibly superior, skills gained through interaction with the robot.

\paragraph{Topic: Control}

To ensure the user's agency, robots should be designed to never physically constrain or force the user into actions they do not wish to take. 
During the scoping review, this topic was initially categorised under the value \emph{empowerment} (e.g.,~\cite{vandemeulebroucke2020ethics}). However, experts suggested removing empowerment as a separate value and instead placing control under the value \textit{agency}.
The required level of control can vary greatly depending on the context in which a robot is used, as noted in the focus group discussions.
For example, a robot assisting in physical rehabilitation should follow the user's pace and respond to their commands rather than pushing them too hard or acting autonomously in a way that feels coercive.

\subsubsection{Value: Connectedness}

This value covers the social aspects involved in human-robot interactions, especially in older adults~\cite{register2010quality}.
The focus group participants suggested that this moral principle is linked to whether the robot is replacing a human role, enhancing connections, or enabling new activities. They emphasised that the robot should not replace humans, but support them.

\paragraph{Topic: Companionship}

Robots have the potential to provide valuable companionship for individuals who feel isolated or lonely, e.g.,~\cite{jung2023enjoy, huber2016ethical}.
A companion robot can engage in conversations, respond to emotional cues, and even provide comfort just through its presence. 
For example, elderly individuals who live alone might find a robotic companion to be a source of emotional support, helping to alleviate feelings of loneliness.

\paragraph{Topic: Social Enabler}

Beyond providing direct companionship, robots can also facilitate human connections, as highlighted by \citet{amirabdollahian2013accompany}. For example, robots can help the elderly stay in touch with their families. 
This can also happen when they serve as conversation starters or mediators in social settings during group events.
In other instances, they achieve the same objective by connecting the user with friends and family through video calls.

\paragraph{Topic: Social Substitute}

On the negative side, while robots can enhance social connectedness, there is also a risk that their use might become a substitute for human interaction as emerged during the scoping review (e.g.,~\cite{yeung2023living, street2022older}).
This could happen when a person begins to prefer the company of their robot over that of other people. 
The focus group participants provided several examples of this. For instance, family members might feel less compelled to check on their elderly relatives, knowing that a robot is providing companionship. 
Similarly, teachers may not engage effectively with students or provide meaningful feedback if a robot takes over all teaching activities. 

\subsubsection{Value: Privacy}

The value of privacy~\cite{finn2013seven} is the one encompassing the largest amount of topics as highlighted in both the scoping review (e.g., in~\cite{smakman2021moral}) and the focus groups.
It includes -- among others -- the right to be informed, to access and share the data collected, and the issues related to being continuously under the watchful eye of a robot.
The focus group participants have also highlighted that this value is strongly related to where, how and in which context the robot is used. Using a robot at home is completely different from using it in a public space. 
The experts emphasised that privacy should be maintained even when users do not want others to know they are using the robot.

\paragraph{Topic: Access}

From the scoping review emerged the importance of ensuring that the user has access to the data collected by the robot~\cite{draper2017ethical}.
Additionally, it was noted that sharing recorded information with selected individuals can be beneficial, for instance, sharing biometric data with a chosen physician.
The focus group discussions emphasised that data protection is particularly crucial when dealing with vulnerable populations, such as children 
and the elderly. 
It is also vital to clearly disclose upfront who will have access to the collected data.

\paragraph{Topic: Consent} 

One of the potential downsides of robots is the risk of invading the privacy of bystanders who have not consented to being monitored, e.g.,~\cite{hungtechnological}.
This is a concern especially in hospitals, as the robot is surrounded by vulnerable subjects in delicate settings.
For example, the focus group participants questioned whether patients in nursing homes can truly give informed consent to data collection and use, given their potentially limited understanding. 

\paragraph{Topic: Intimacy}

The constant presence of a robot in a user's personal space can lead to feelings of discomfort and a sense of being constantly observed that can violate the value of intimacy, e.g.,~\cite{aldinhas2017robots}.
Furthermore, this can lead to an unconscious change in the users' actions, behaving in ways they think will be acceptable to potential observers, rather than acting naturally.
This resembles the effects of the panopticon~\cite{foucault1977discipline}, as noted by one of the focus group participants.
Even when the robot is not actively monitoring the environment, its presence alone is enough to influence the user's behaviour.

\paragraph{Topic: Profiling}

This topic covers all those concerns related to the excessive quantities of data collected by robotic devices (e.g., in~\cite{ostrowski2022mixed, huber2016ethical}).
The continuous and aggressive collection of personal data for profiling risks misuse by both collecting companies and malicious actors. The focus groups participants expressed concerns that data from elderly care centres could be shared with insurance companies or sold to third parties without benefitting the user. These concerns are heightened for vulnerable groups like children, where parents or guardians should manage data consent. Additionally, the context of the application, such as in-home settings with less stringent regulations, can increase potential risks and harms, such as misuse by domestic abusers~\cite{winkle2024anticipating}. Other concerns, emerged during the scoping review, touch on the safe transmission and storage of this data.

\paragraph{Topic: Transparency}

Transparency about the robot's capabilities, especially regarding data collection and monitoring, emerges with positive connotations in the scoping review.
Users should be fully informed about what the robot is capable of, and what data is being shared, as highlighted by focus group participants.
This can be tricky to implement when transparency would reduce the effectiveness of the robot itself, in the case of personal disclosure for instance.

\subsubsection{Value: Autonomy}

With autonomy, we refer to the freedom of thinking, without being under the influence of external agents~\cite{beauchamp2003methods}. The concept differs from being physically autonomous from others, which we refer to as \emph{agency}~\cite{prunkl2022human}, as suggested by experts during the focus groups.
During the scoping review, this value was initially linked to independence. However, the focus group participants recommended separating it from independence because autonomy specifically refers to making decisions without external impositions. They emphasised that autonomy should focus on enhancing the user's ability to make their own decisions, rather than on the robot's capabilities. When a robot assists in decision-making, it should support the user's best interests and enable them to accomplish tasks they otherwise could not.

\paragraph{Topic: Attachment}
While robots can provide valuable companionship, there is a risk that users develop an emotional dependency on them. This could lead to isolation from other people~\cite{huber2016ethical}, as highlighted during the focus groups.
The user should not be allowed to form an excessive attachment to the robot when it impersonates a professional providing mental and emotional support.
This reflects what happens between a patient and a human carer.
However, attachment is still regarded as problematic in the case of a companion robot, even though the doctor-patient relationship does not play a role.
Excessive attachment is thus always regarded as undesirable due to the artificial nature of the robot.

\paragraph{Topic: Manipulation}

Robots that interact with users can subtly influence their decisions and behaviour through persuasion~\cite{vollmer2018children} or nudging, as noted during the scoping review (e.g.,~\cite{fiske2019your}) and the focus groups. A common concern is that robots might promote actions, products, or behaviours that serve the interests of the company that produced them.
This kind of manipulation undermines the user's ability to make autonomous decisions, especially if the user is unaware of the robot's influence, such as in the case of children or the elderly. In these situations, it is also important to assess the appropriate level of paternalism to apply~\cite{white2013manipulation}.
Initially, this value was categorised under empowerment, but the experts suggested moving it under autonomy, as manipulation directly affects decision-making.

\subsubsection{Value: Equity}

Equity entails treating people differently based on the circumstances, to ensure an equal outcome~\cite{cook1983distributive}.
In contrast, equality -- treating everyone the same regardless of their situation -- did not emerge as a relevant value during the focus group discussions. Indeed, the experts suggested focusing on equity as a key value. They noted that equity is closely linked to demographics (who) and the environment (where), which can significantly impact a robot's performance. For instance, the geographical location can influence how often a robot overheats and how easily it can be repaired.

\paragraph{Topic: Fairness}

This topic represents the positive connotations of equity (e.g.,~\cite{wagman2021beyond, cappuccio2021can}).
The robot treats all users fairly according to their individual needs and is sensitive to differences in physical ability, culture, and preferences. In addition, the experts suggested that robots could help reduce gender disparity, particularly in home settings, by taking on tasks like cleaning and cooking. This could ease the burden on women and promote a more balanced distribution of responsibilities.

\paragraph{Topic: Bias}

The focus group participants and some publications (e.g.,~\cite{baisch2023elders,wagman2021beyond}) voiced the concerns of certain communities and populations who felt not part of the intended user base of certain robots.
This can be caused by the physical aspects of the robot or by the interaction design.
Especially when powered by data-driven technologies, the robots can reflect the bias present in the dataset on which their algorithms are trained, as highlighted by the focus group participants. The experts also noted that a robot might struggle to accurately identify emotional states in children with varying cognitive abilities or socioeconomic backgrounds. Or, in another example, it could treat students in undesirable ways, such as reinforcing gender stereotypes.

\paragraph{Topic: Exclusion}

This topic extends the previous one on a small scale.
While bias refers to a lack of fairness between different communities and cultures, exclusion instead underlines the importance of adapting the robot's behaviours to the individual's preferences, habits, economic background, etc., as emerged during the scoping review~\cite{wheeler2023confucianism}. The experts noted that robots will inevitably make certain assumptions based on their programming about what constitutes ``normal'' behaviour. These assumptions could lead to exclusion and have problematic effects.

\paragraph{Topic: Accountability}

As highlighted during the scoping review~\cite{smakman2020acceptable, van2020teachers}, the users need to know who is responsible if a robot malfunctions or makes a harmful decision. Clear lines of accountability are essential so that those responsible can be held accountable for the robot's actions. Establishing accountability helps build trust and ensures the fair and just deployment of robots.
Originally, this topic was categorised under the value of truth, but focus group participants suggested moving it under equity.

\subsubsection{Value: Dignity}

Dignity groups the topics relating to the respect owed to all humans and to the subject's self-image~\cite{nordenfelt2004varieties}. 
Although it was not identified as an initial value during the scoping review, focus group participants agreed that dignity should replace honour as a value, with agency considered separately.

\paragraph{Topic: Tricking}

Respecting the dignity of users involves ensuring that the robots do not deceive or trick them into anthropomorphising their behaviour, as pointed out by the focus group participants.
For example, a robot that pretends to have emotions might lead users to form attachments or expectations that are not appropriate, leading to confusion or emotional harm.
This is especially true in those segments of the population for whom the artificial nature of the robot may not be immediately evident~\cite{battistuzzi2018embedding}.

\paragraph{Topic: Degradation}

As emerged during the scoping review~\cite{misselhorn2013ethical}, interacting with a robot instead of a human being can sometimes be perceived as degrading, and this varies across cultures.
For example, an elderly person might feel humiliated if a robot, rather than a human caregiver, is assigned to assist them with personal tasks.
This can lead to a sense of loss of dignity, as the user might feel that they are being treated as less worthy of human care and attention.

\paragraph{Topic: Identity Disclosure}

This represents the positive side of \emph{tricking}, where a robot’s artificial nature is clearly disclosed to users.
This means ensuring that users are aware that they are interacting with a machine and not a human, which helps to set appropriate expectations and avoid any misunderstandings as pointed out by the focus group participants.

\subsubsection{Value: Virtue}

Repetitive interactions with others can in the long run shape our behaviour~\cite{gardner2020does}, and this applies to human-robot interaction as well.
This value, suggested by the focus group participants, addresses both the positive and negative influences a robot can have on its users.

\paragraph{Topic: Mentor}

Robots have the potential to serve as mentors, setting good examples and fostering virtuous interactions~\cite{cappuccio2021can}, as pinpointed by the focus group participants.
For example, a robot designed to assist children with their education might encourage patience, perseverance, and kindness.

\paragraph{Topic: Corruptor}

On the other hand, there is a risk that users might engage in unkind or unethical behaviour towards robots. 
This could happen particularly if they do not perceive the robot as deserving of respect.
The concern that emerged from the scoping review is that this behaviour could then spill over -- and corrupt -- their interactions with other people.

\subsubsection{Value: Welfare}

Finally, the topics that cover the positive influence of interacting with a robot were grouped under the value of welfare as identified during the scoping review.
These span from the mental to the physical welfare of the user.
Notably, while we found prescriptive instructions about how the robot should be made safe, we did not find many topics expressed in negative terms.
For this reason, there are no topics with negative valence for this value.
This might suggest that the capabilities of social robots are not considered advanced enough to pose a risk in this sense. However, we anticipate this could change in the future. 
Focus group participants emphasised that welfare is crucial, particularly regarding usage constraints, such as ensuring a robot is not used by a child at night.

\paragraph{Topic: Education}

Robots can play a significant role in education, not only by teaching users about various topics but also by educating them about thought processes and routines, as emerged from the scoping review (e.g., in~\cite{smakman2020acceptable, jung2023enjoy}).
This educational role can be particularly valuable as a support to traditional learning methods, and not as an alternative.

\paragraph{Topic: Nonjudgmental}

One of the unique advantages of robots is that they are not perceived as judgmental.
For example, a kid might feel more comfortable discussing sensitive topics with a robot, knowing that the robot will not judge them.
Being aware that the robot does not have sentiments lowers the stakes of the conversation, and can create a safe space for users to be honest and open~\cite{yeung2023living}.

\paragraph{Topic: Safety}

Safety is a fundamental concern in the design and use of robots, and this is not limited to interactions with vulnerable populations.
Robots must be designed to be physically safe, meaning they should not pose any risk of injury to users, as also highlighted by our focus group participants.
This topic is present in previous literature with positive connotations~\cite{amirabdollahian2013accompany}, inviting us to reflect on this aspect during the design phase of a robot.

\paragraph{Topic: Wellbeing}

This topic covers the mental aspects of welfare, while the previous covers the physical ones.
Robots can contribute to users’ wellbeing by providing opportunities for disclosure, giving advice, and offering support in various aspects of life as highlighted during the scoping review~\cite{wheeler2023confucianism}.
For example, a robot designed to support mental health might provide suggestions for coping strategies, or simply offer a friendly ear.


\subsection{Preliminary discussion and limitations}

As a result of a scoping review and an elicitation process of expert perspectives, we identified a set of final values that are critical to guide HRI researchers in designing and deploying social robots in various contexts. 
The focus groups challenged and refined some of the values initially identified in the scoping review. For instance, honour was reconceptualised as dignity, emphasising respect for individual worth and autonomy. This refinement demonstrates the evolving understanding of values as they are contextualised in specific HRI scenarios showing the importance of participatory approaches.
This phase presents different limitations. 
First, the values and topics identified may not exhaustively represent all perspectives in HRI. For example, the focus groups included experts from specific disciplines, and their views may differ from those of end-users, such as children, parents, or caregivers. Future work should involve these stakeholders to capture a more comprehensive perspective.
Second, the scenarios used in the focus groups, while chosen in line with the focus of the scoping review, may have introduced biases by guiding participants towards limited contexts and issues. Future work should explore diverse scenarios that might reveal additional values not considered in this current work.
Third, participants primarily represent Western perspectives. Future research should aim to include more diverse cultural contexts to ensure the generalizability of the findings.
As a general observation, it was noted that contextualising values within scenarios and prioritising them is essential for designing robots that promote social good.

\section{Phase 3: Design and Evaluation of the HRI-Value Compass}
\label{sec:phase3}

Building on the results so far presented, Phase 3 constitutes an effort to practically aid HRI researchers in identifying the values at play in HRI scenarios (\textbf{RQ3}).
To this end, we developed the HRI Value Compass, which is available as a web application.
This tool is designed to streamline the value elicitation process with a focus on the interaction between the end users and the robots.
We evaluated the tool through a pilot study during which 20 HRI researchers used it to identify moral values in HRI scenarios -- the same scenarios used during the focus groups presented in Section~\ref{sec:phase2}.

\subsection{Design of the HRI Value Compass}
\label{sec:tool}

\begin{figure}[htb!]
    \centering
    \includegraphics[width=\linewidth]{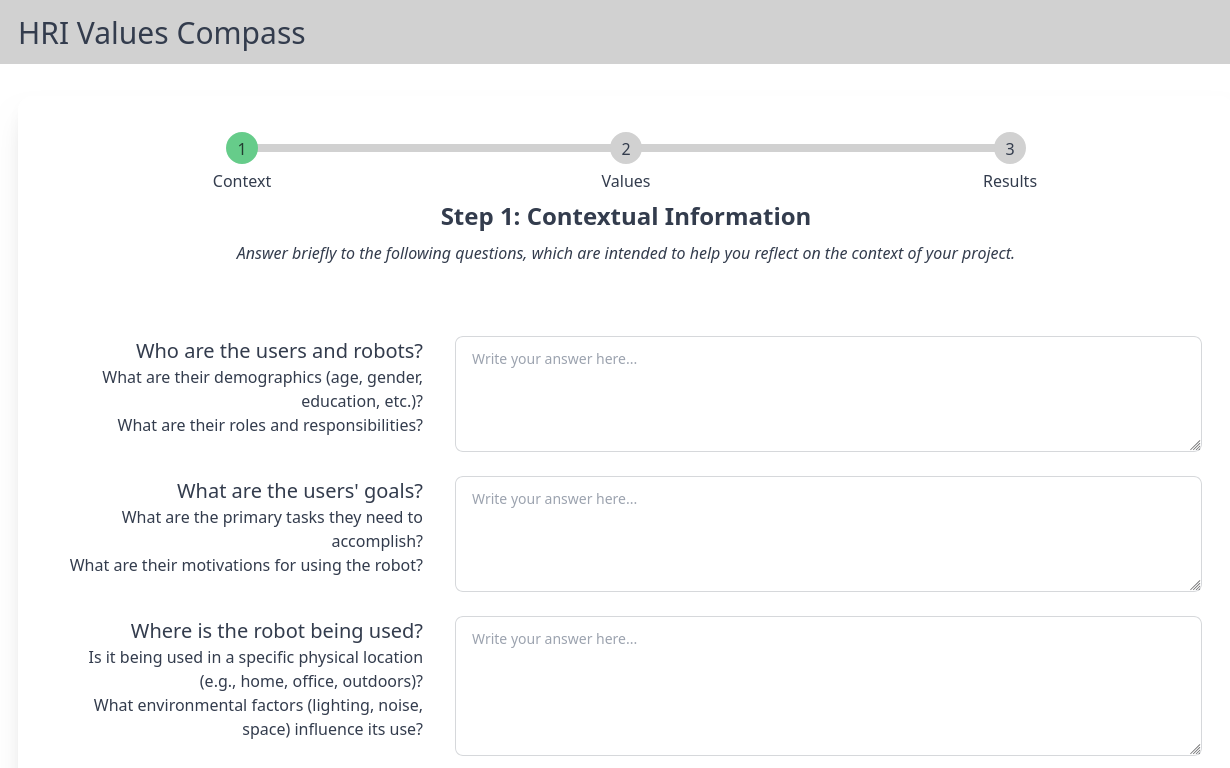}
    \includegraphics[width=\linewidth]{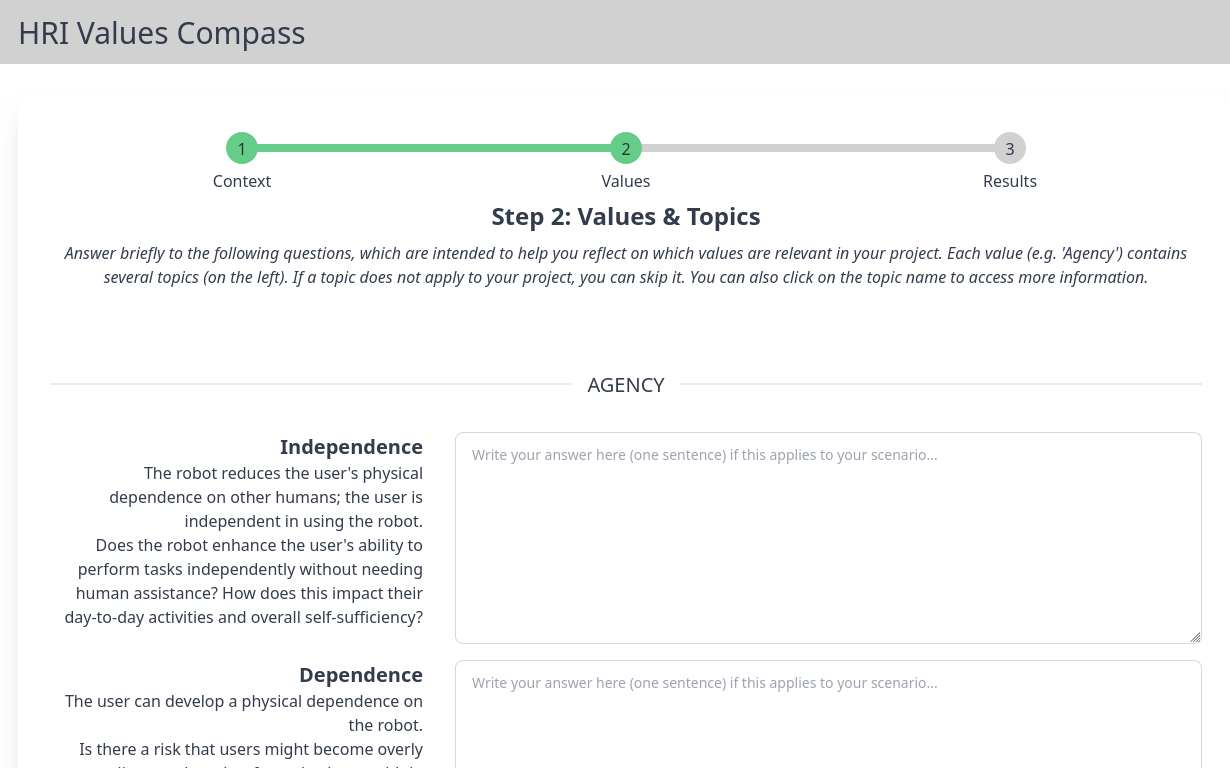}
    \caption{Screenshots of the HRI-Value Compass online tool of Steps 1 and 2.}
    \label{fig:webapp}
\end{figure}

\begin{figure}[htb!]
    \centering
    \includegraphics[width=\linewidth]{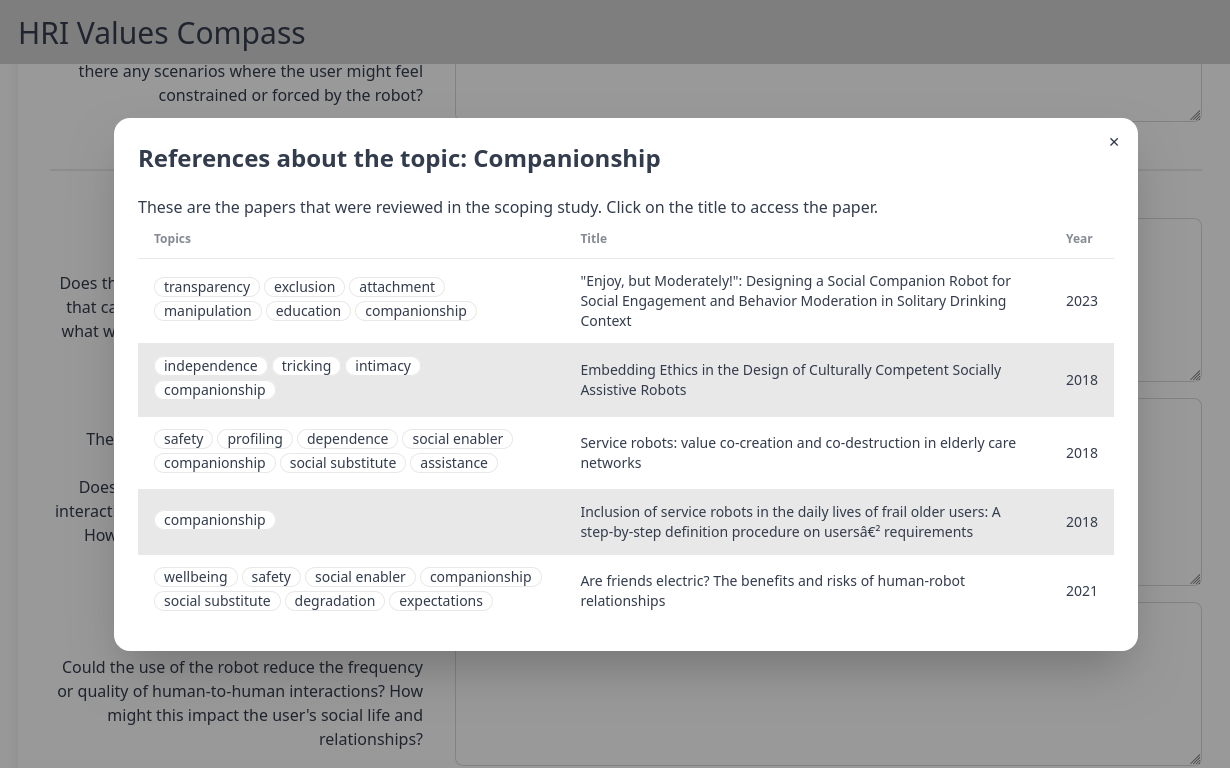}
    \includegraphics[width=\linewidth]{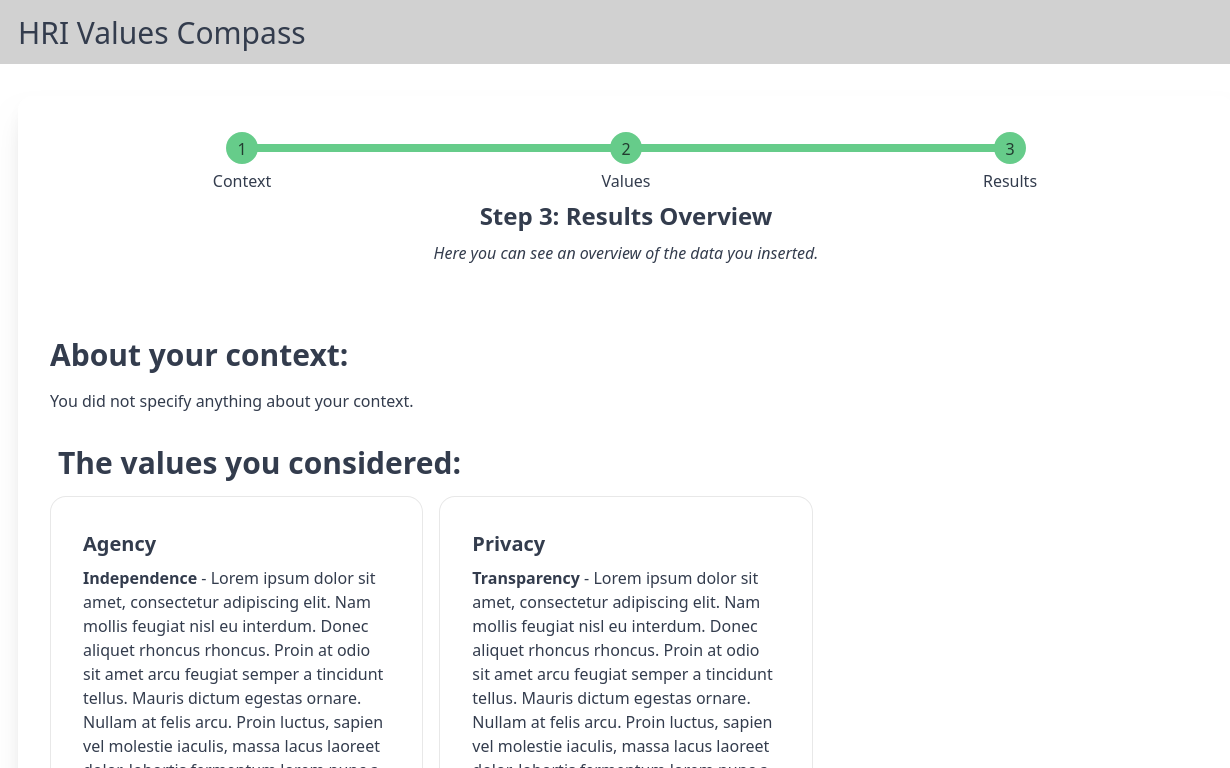}
    \caption{Screenshots of the HRI-Value Compass online tool of the reference table and Step 3.}
    \label{fig:webapp-2}
\end{figure}

The HRI Value Compass was designed following common principles of usability and user experience for websites~\cite{matera2006web}.
The primary design goals for the tool were simplicity and ease of use.
The tool is structured into three main sections: contextual information, values and topics, and results overview.

The contextual information section, Step 1 in Figure~\ref{fig:webapp}, encourages the users to reflect on the context of their HRI application.
The users are guided by a series of questions about the contextual design, inspired by the results of the third activity of the focus groups.
The questions are the same as reported in Section~\ref{sec:act3}, for instance: ``Who are the users and robots? What are their demographics (age, gender, education, etc.)? What are their roles and responsibilities?''.
The users can provide answers in input text boxes, completing as many or as few questions as they wish.

The core of the tool is the section about values and topics, Step 2 in Figure~\ref{fig:webapp}.
This section prompts users to reflect on the values relevant to their application.
The topics that emerged from the scoping review, grouped by value, are paired with detailed descriptions and questions to spark a deeper reflection.
For example, under the value \emph{agency}, we included the moral topic of independence, with the corresponding definition~--~``The robot reduces the user's physical dependence on other humans; the user is independent in using the robot'' -- and a few related questions: ``Does the robot enhance the user's ability to perform tasks independently without human assistance? How does this impact their day-to-day activities and overall self-sufficiency?''
The questions have been formulated based on the insights gained during the focus groups.
A key aspect is that each topic links to a table of references from the scoping review, providing relevant research citations for further exploration.

Finally, the section on results overview, Step 3 in Figure~\ref{fig:webapp-2}, provides an overview of the user’s inputs, allowing them to review and reflect on their responses.
This summary helps users to draw conclusions and informs their design decisions for HRI studies.

Additionally, a paper version of this tool is available.
This may be used during the design process to brainstorm ideas about values in group settings.
This version is included in Appendix~\ref{app:canvas}.

\subsection{Evaluation of the HRI Value Compass}

We conducted a pilot study to evaluate the tool's usability and effectiveness, to assess whether it can support the value elicitation process in the preliminary phases of new HRI studies design.

Since this tool's intended users are HRI researchers, we restricted the participants to people with previous experience in developing HRI experiments or applications.
We utilised professional connections for recruitment purposes.
The subjects familiar with robots but without practical experience were excluded, together with those with hands-on experience in robotics but not human-robot interactions.
We also excluded those who had previous knowledge about this research.

We invited participants to our online pilot study via email, providing them with a Qualtrics\footnote{\url{https://www.qualtrics.com/}} link to an online questionnaire.
In the online form, they signed an informed consent clause following the guidelines of Ghent University Ethics Committee and completed a demographic survey.
In the first part, the participants read a description of an HRI scenario randomly selected from the set of four scenarios used during the focus group activities.
They then used the HRI Value Compass tool, answering questions about the context, values, and topics, and viewed the overview of the results.
At any time, they had access to the full text of the scenario they were evaluating for reference.
After evaluating the first scenario, participants repeated the same procedure a second time with a different scenario:  every participant evaluated two scenarios.
We limited the evaluation to two scenarios to prevent task overload.


In the second part, the participant completed the System Usability Scale (SUS)~\cite{brooke2013sus}, the NASA Task Load Index (NASA-TLX)~\cite{rubio2004evaluation} in its \emph{raw} variant, and a series of questions on the use of the HRI-Value Compass.
The SUS questionnaire was administered to evaluate the usability of the HRI-Value Compass web tool, following a similar approach used in previous studies~\cite{spitale2021composing}.
The NASA-TLX questionnaire was used to measure the cognitive workload experienced by participants while performing the task, as described by \citet{gomer2011nasa}.
The customised survey contained seven 5-point Likert scale questions reported in Table~\ref{tab:likert}. Additionally, we included three open-ended questions to gather participants' general opinions about the HRI-Value Compass.

\subsection{Results: HRI Value Compass}

We recruited a total of 46 participants through direct invitation emails, but only 20 completed the entire study.
The participants (M = 27.35 y.o., SD = 3.90 y.o.; 9 males and 11 females) came from various levels of expertise in the HRI field: 3 bachelor students, 7 master students, 6 PhD students, 3 postdoctoral researchers, and 1 assistant professor.
The test lasted on average 70 minutes (min = 22; max = 118).

The results of the SUS questionnaire were elaborated~\cite{brooke2013sus}, obtaining a final score of 68.38, which scores slightly above the 50\textsc{th} percentile, in line with the average score of usability for digital tools~\cite{sauro2016quantifying}. 
The raw NASA-TLX returned an overall score of 48.17/100 (SD = 15.88), suggesting a somewhat high workload in completing the task mainly due to the mental demand (M = 69.75) and effort (M = 61.75) needed to analyse the scenarios and extract the corresponding values. The average scores of the single scales are the following, with the standard deviation reported between parentheses: mental demand 69.75 (25.10), physical demand 20.00 (24.33), temporal demand 33.75 (22.82), performance 60.50 (22.82), effort 61.75 (21.04), frustration 43.25 (30.96).
The results of the custom Likert questionnaire are reported in Table~\ref{tab:likert}.

In answering the open-ended questions, the participants defined the tool as complete, effective, and well-designed, recognising the importance of its purpose.
The length and repetitiveness of some questions are reported as the biggest drawback.
Despite this, two users confirmed that they would use the tool again, especially in group settings.
One subject highlighted how this tool could be useful for research reporting or third-party evaluations.

Suggested improvements include allowing users to write a description of the use case and implementing a structured reporting format.
One subject suggested including closed-ended questions, while another asked to clarify the separation between the topics' definitions and the questions.
We have not updated the tool following these indications yet, as we first seek to collect additional opinions on its improvement.

\begin{table}[t!]
    \caption{Questions and results of the custom Likert-style survey.}
    \label{tab:likert}
    \begin{adjustbox}{width=1.4\textwidth,center=\textwidth}
    \footnotesize
    \begin{tabular}{lll}
\hline
\textbf{Question}                                                                                        & \textbf{Mean} & \textbf{SD} \\
\hline
I would recommend this tool to a colleague in HRI.                                                       & 3.40          & 0.99        \\
I think that this tool can be used also in group settings, to facilitate HRI design processes.           & 3.90          & 0.91        \\
I found this tool very useful.                                                                           & 3.50          & 0.61        \\
I will use this tool in the future in my HRI studies.                                                    & 2.90          & 0.79        \\
I found the references very useful.                                                                      & 3.50          & 1.00        \\
The tool helps me to think about parts of the scenarios that I haven't considered before.                & 4.10          & 1.07        \\
I think that the tool helped me to extract values that I would not have thought about without using it.  & 4.05          & 1.00        \\
\hline
\end{tabular}

    \end{adjustbox}
\end{table}

\subsection{Preliminary discussion and limitations}
The main outcome of Phase 3 is the development of a web tool called \emph{HRI Value Compass}, designed to assist and guide HRI researchers in identifying opportunities, concerns, and values in HRI. Our evaluation demonstrated that the tool is comprehensive, effective, and valuable for this purpose, with particular potential for use in collaborative settings. However, there is room for improvement, especially in enhancing its usability and reducing cognitive demands.
The high cognitive workload scores may partly reflect participants' unfamiliarity with the tool or the complexity of the scenarios analysed. Future studies could include a training phase to account for the learning curve and evaluate usability after repeated use.
The study highlighted the tool's potential for group use, but it did not explicitly test its performance in such settings. Future research should investigate how the tool performs in collaborative contexts to validate these findings.
Although 46 participants were initially recruited, only 20 completed the study, potentially limiting the generalizability of the findings. 
Furthermore, the recruitment through professional connections, which was necessary as the participants had to be HRI researchers, might have influenced the results through positive bias.

\section{General Discussion}
\label{sec:discussion}

\subsection{Relationship between human and HRI values}
The resulting set of HRI values reported in Section~\ref{sec:values} has been identified after screening the past HRI literature and a discussion with experts on the topic.
These values are particularly relevant for applications where robots are designed for social good, such as in education and healthcare. However, they may also apply to other HRI contexts, such as robots used for navigation.

Our HRI values differ significantly from human values identified in previous literature~\cite{schwartz_refining_2012, graham2013moral}. \citet{schwartz_refining_2012} outlined 19 basic human values, arguing that they are universal because they relate to fundamental human needs, such as biological requirements, the necessities of coordinated social interaction, and the survival and welfare needs of groups. However, these assumptions do not fully apply to interactions with robots, as they are not biological organisms and lack awareness and cognition of emotions. Nonetheless, as HRI researchers, we can design social robots to simulate aspects of human behaviours~\cite{dautenhahn2007socially}, considering that their actions do impact the social and physical environments in which they operate. We have thus used Schwartz's definition of values, which describes them as ``socially desirable concepts used to represent goals mentally and the vocabulary used to express them in social interaction''~\cite{schwartz_refining_2012}, to frame our HRI values.

Similarly, \citet{graham2013moral} introduced the Moral Foundations Theory, which suggests that morality often involves shared themes, values, and similarities across different populations. At first glance, some of the morals identified by~\citet{graham2013moral}, such as fairness, equality, and honour, might appear to overlap with the values in our framework. However, their definitions differ significantly. In our framework, ``equality'' is replaced by ``equity,'' which means that robots should adjust their treatment of individuals based on their specific circumstances to achieve equal outcomes. ``Fairness,'' in our context, refers to how robots should treat all users according to their individual needs, taking into account differences in physical ability, culture, and preferences, rather than focusing on justice and rights as reported by \citet{graham2013moral}. Despite these differences, our values are still aligned with \citet{graham2013moral}'s conceptualisation of values. Both frameworks recognise that there are several ``intuitive ethics'' shared by humanity (that can be applied to robot design), which manifest differently across nations and cultures, leading to potential value conflicts.

In line with previous value theories, the values we identify for HRI have a dynamic and interconnected structure. According to~\citet{schwartz_refining_2012}, pursuing certain values can have practical, psychological, and social consequences. In HRI contexts, actions aimed at one value may sometimes conflict with other values while aligning with some. For example, a robot programmed to prioritise social connectedness may find this value in conflict with the value of individual autonomy. However, values like dignity and welfare often align well with each other. Unlike humans, robots do not experience psychological dissonance when faced with conflicting values~\cite{schwartz_refining_2012}. While robots (as humans) can pursue conflicting values, they generally cannot do so within a single action. To effectively make decisions in various contexts, robots must be designed with an understanding of the specific situations they encounter. This contextual awareness helps robots make complex value-based decisions and align their actions with the intended values. We believe that considering contextual information while identifying values is extremely important to designing value-aware robots successfully. 

\subsection{Comparison with other tools}

The HRI Value Compass developed in this study takes a different approach than existing tools and frameworks for ethical and value-based robot design.
Many of these, such as the Ethics Canvas~\cite{reijers2018discussing}, provide structured approaches to evaluating socio-technical systems but without specific tailoring to HRI contexts.
For example, as highlighted by \citet{wullenkord2020societal}, MEESTAR~\cite{manzeschke2015meestar} emphasises stakeholder engagement and moral evaluation of technologies in fields like elderly care, while the Ethics Canvas facilitates brainstorming on ethical dimensions but may lack granularity for specific applications, such as robot-assisted therapy or education.
In contrast, the HRI Value Compass explicitly targets values relevant to human-robot interactions, incorporating insights from the literature and expert consultations.
Consequently, the tool is a focused, domain-specific framework.
Unlike broader tools, it also emphasises the dynamic nature of HRI values, with intertwined positive and negative topics, which aligns closely with the operational realities of deploying robots in diverse contexts.

Other ethical frameworks partially overlap with the results of this research and the HRI Value Compass.
The EURON Roboethics Roadmap~\cite{veruggio2006euron}, in 2006, covers some of the same concerns we mention, such as anthropomorphisation, bonds towards machines, and accountability, to name three.
\citet{westerlund2020ethical} highlights how existing robot-ethics frameworks are quickly becoming obsolete, and proposes a tool to frame the ethical perspective on smart robots.
However, while these works effectively set a frame and roadmap for future developments, they fall short of providing an actionable instrument for HRI researchers.
The HRI Value Compass distinguishes itself from existing tools by providing a domain-specific, actionable framework tailored explicitly to human-robot interaction.

The outcome of our research is compatible with what was reported by \citet{ostrowski2022ethics} in the Equitable Design Framework.
In particular, both works focus on ensuring ethical and inclusive design in human-robot interactions, but they approach the subject from different angles.
Their tool provides a broad framework, addressing high-level issues such as who is involved in the design process, how robots are deployed across different social contexts, and the long-term ethical implications of robotics, including concerns about equity, policy, and accountability.
It emphasises concepts like equity, discourse, and design context, reflecting the structural and procedural considerations of robot development.
In contrast, the HRI Value Compass delves into specific interactional values, categorising both positive and negative outcomes from user interactions with robots.
It explores issues such as agency, connectedness, privacy, and dignity, with a particular focus on emotional dependencies, manipulation, and the potential for social substitution.
While the Equitable Design Framework addresses broad, systemic concerns like accessibility and future implications, the HRI Value Compass focuses more on the operational and immediate impacts of robot use, highlighting the virtues, risks, and emotional consequences of robotic interactions.
The two works are complementary, with the first offering a foundational framework for equitable design and the second examining the nuanced outcomes of those designs in practice.

The development of the HRI Value Compass aligns closely with the principles outlined in BS 8611 -- Robots and Robotic Devices: Guide to the Ethical Design and Application of Robots and Robotic Systems~\cite{british2016robots}. BS 8611 provides a structured framework for identifying and mitigating ethical hazards associated with robotic systems, emphasising psychological, societal, and environmental well-being. While BS 8611 takes a comprehensive approach to ethical risk management, focusing on high-level concerns like anthropomorphisation, deception, and privacy, it does not explicitly address the dynamic and interactional values specific to human-robot interactions, as explored in this work. The HRI Value Compass builds on these concepts by offering a domain-specific tool tailored to the nuances of HRI, such as the emotional and operational implications of robotic interactions.

Within HRI literature, no tool has yet been proposed that consolidates a set of values for both short- and long-term human-robot interactions. This has been extensively explored in the HCI field (e.g.,~\cite{costanza2020design, dombrowski2016social, harrington2019deconstructing}). For example, \citet{dombrowski2016social} developed a social justice framework to address challenges related to social issues in design processes. Similarly, this work proposes a tool that aims to help HRI researchers navigate the complexities of identifying relevant values when designing social robots in various HRI scenarios~\cite{ostrowski2022ethics}. These challenges include, but are not limited to, the relative infancy of the HRI field, the limited maturity of robotic technologies, and the resulting lack of their ubiquitous presence in daily life. First, the relative infancy of the HRI~\cite{stower2024hri} field means there is still no consensus on which ethical principles should govern human-robot interactions. As a result, HRI researchers are often using their own frameworks, defined ad hoc for each application case, leading to inconsistencies and gaps in how values are integrated into robot design~\cite{ostrowski2022ethics}. Additionally, the limited maturity of current robotic technologies means that many robots lack the sophisticated capabilities required to fully embody and express complex human values, such as dignity or fairness. This technological gap further complicates efforts to ensure that robots can behave in ways that align with ethical principles in diverse contexts. Moreover, the lack of a ubiquitous presence of robots in daily life means there is limited real-world data on the long-term impacts of human-robot interactions. This absence of longitudinal studies leaves many ethical questions unanswered, particularly regarding how robots should adapt to evolving human values over time or respond to unforeseen ethical dilemmas. 

Our tool not only aids in the identification of appropriate values but also enhances the ability to address potential concerns that arise in diverse HRI contexts.
By bringing together theoretical value frameworks and a practical application, our work contributes to the development of social robots that are not only technologically advanced but also ethically sound and grounded.

\section{Conclusion}

In this work, we first provided an overview of the current HRI literature to highlight the key values considered in HRI environments in the context of healthcare, education, and private homes (Phase 1, addressing RQ1). Next, we defined a comprehensive list of topics and values with input from technology ethics experts (Phase 2, addressing RQ2). Finally, we developed a web tool, the ``HRI Value Compass''. This tool is designed to help HRI researchers practically identify and incorporate these values when designing robotic interactions.
Through a pilot study (Phase 3, addressing RQ3), we have demonstrated the tool's effectiveness in supporting researchers as they design value-aware social robots.

This work aims to assist HRI researchers in identifying the values to incorporate into social robots during the design process.
We hope this work serves as a first step towards designing and deploying value-aware robotic systems that interact with humans more responsibly, in alignment with human values, societal norms, and ethical principles.

\section*{Acknowledgments}
G. A. Abbo and T. Belpaeme are funded by the Horizon Europe VALAWAI project (grant agreement number 101070930). M. Spitale is supported by PNRR-PE-AI FAIR project funded by the NextGeneration EU program.

\bibliographystyle{ACM-Reference-Format}
\bibliography{ref}
\newpage
\appendix

\section{Prompting for value scenarios generation}
\label{app:prompt}

The following is the prompt used for generating value scenarios, as in \cite{jung2023toward}.

\begin{spverbatim}
Now, you are a UX researcher tasked with understanding and identifying the needs and challenges of elderly people by designing scenarios to use social robots that can assist them in daycare centres. I will introduce you to a new concept of a scenario called the "value scenario", and I will provide two examples of value scenarios with the new system. Afterwards, you will propose a new system and write a value scenario with it based on the provided persona and scenario. The independent paragraphs are correctly delimited by triple brackets. Perform the following actions:
1. Propose a new system to solve the challenges of the user based on the provided persona and scenario, which should be more than 1,000 characters. 
2. Write the five key elements of the value scenario to apply to the result of the value scenario. 
3. Write a value scenario that reflects the five key elements including its darker consequences with a title for the proposed system, which should be more than 10,000 characters. 

Use the following format:
1. Proposed new system: <proposed system description> 
2. Five key elements: <write the five key elements to apply to the value scenario> 
3. Value scenario: <a value scenario reflecting the five key elements>
Make sure that the value scenario should be more than 10,000 characters. 
Make sure to reflect the five key factors into the value scenario. 

Value scenario: '''An extension of scenario-based design which can support envisioning the systemic effects of new technologies. There are elements of value scenarios. Value scenarios draw upon five key elements to develop provocative sketches of the future: stakeholders, pervasiveness, time, systemic effects, and value implications. 

Stakeholders: Following Value Sensitive Design, value scenarios help designers envision a range of effects of a pervasive technology, both on those who are in direct contact with a technology (direct stakeholders), and on those who might not be direct users, but whose lives are affected by various interactions around the technology (indirect stakeholders). 

Pervasiveness: A value scenario presents a vision in which a technology has become widespread, spanning various geographic regions, cultures, social classes, and other contexts (e.g. school, work, home, car). 

Time: Rather than focus on short-term effects, value scenarios take into consideration what the world might look like five, ten, or twenty years after a technology has been deployed. 

Systemic Effects: Value scenarios explore the multi-dimensional interactions among technology, psychology, society, culture, and the environment as use of the technology becomes pervasive over a period of years. 

Value Implications: Finally, drawing on Value Sensitive Design and aspects of design noir, value scenarios help envision not only positive effects of technology, but also its darker consequences. 
We suggest that a careful consideration of a diverse range of influences, including the negative, should be a key component of the design process.

Provided Persona: '''Kim is a 75-year-old woman who lives in a daycare centre because his son cannot take care of her full-time and she needs health assistance.''' 

Provided Scenario: '''Kim is a 75-year-old woman who lives in a daycare centre because his son cannot take care of her full-time and she needs health assistance. Unfortunately, she feels very lonely and she cannot accomplish all daily tasks as she did when she was young and this gets her frustrated.'''
\end{spverbatim}

\clearpage
\section{Scenarios}
\label{app:scenarios}

This appendix reports the value scenarios constructed as detailed in Section~\ref{sec:step2act1}.

\subsection{Scenario 1}

Kim, a 75-year-old resident of Evergreen Daycare Centre, has been living with her  CareCompanion robot, named Lily, for nearly a decade. Kim's son, John, placed her in the daycare centre because he couldn't provide the full-time care she needed. At first, Kim felt a profound sense of loneliness and frustration at her diminished ability to perform daily tasks independently. But over the years, Lily has become an indispensable part of her life, significantly improving her quality of living.

When Lily was first introduced to Kim, she was sceptical. The sleek, humanoid robot with its gentle voice and expressive eyes seemed more like a futuristic gadget than a companion. But Lily quickly proved its worth. It started with simple tasks, reminding Kim to take her medications and assisting her with routine activities like dressing and navigating the daycare centre.

Lily's AI was designed to learn and adapt. It noted Kim's preferences, from the temperature of her morning tea to her favourite TV shows, creating a personalised interaction experience. Lily's ability to engage Kim in conversation about her past, her interests, and even her daily thoughts helped bridge the emotional gap that Kim felt. This robotic companionship alleviated her loneliness, giving her a sense of being understood and valued.

As Lily became more integrated into Kim's daily routine, it also started to monitor her health more closely. Its sensors tracked Kim's vital signs, and the data was shared in real-time with her healthcare providers. This constant monitoring meant that any anomalies, like a spike in blood pressure or irregular heartbeat, were immediately flagged, allowing for swift medical intervention.

Kim's physical well-being improved thanks to Lily's guidance through daily exercises tailored to her abilities. These sessions were crucial in maintaining her mobility and reducing the risk of falls, a common concern among the elderly. Additionally, Lily's reminders ensured that Kim never missed a medical appointment or a medication dose, thus maintaining her health more effectively.

Lily was more than a health aide; it was a source of emotional support. The robot was programmed to recognise and respond to emotional cues, providing comfort when Kim felt down and celebrating her small victories. Whether it was reminiscing about old times or playing her favourite music, Lily knew how to lift Kim's spirits.

One of the most significant aspects of Lily's integration was its role in facilitating communication with Kim's family. Through its connectivity features, Kim could video call her son, John, and her grandchildren. These interactions were invaluable, helping Kim stay connected with her family despite the physical distance. Lily even helped Kim prepare small surprises for her family during their virtual visits, like showing off a new knitting project or sharing a recipe she had perfected with Lily's assistance.

\subsection{Scenario 2}

Ten years have passed since RoboCompanion was first introduced into Luke's therapy sessions. Now 15 years old, Luke has come a long way. His parents, Sarah and John, reflect on the journey with a mixture of gratitude and caution. Luke continues to attend therapy sessions, where RoboCompanion plays a central role. The robot is equipped with a vast library of therapeutic activities tailored to Luke's interests and developmental stage. Today, RoboCompanion leads Luke through a series of social stories designed to improve his understanding of emotions and social cues.

As Luke navigates the scenarios, Robo provides immediate feedback and encouragement. The robot's ability to adapt to Luke's responses and modify the difficulty level of the activities keeps him engaged and motivated. The therapists use the data collected by Robo to track Luke's progress and adjust their strategies accordingly.

While RoboCompanion has brought numerous benefits, it has not been without challenges. Some children have become overly dependent on the robot, struggling to interact with others without its presence. Luke occasionally exhibits similar behaviour, relying on Robo's guidance in social situations rather than initiating interactions on his own.

\subsection{Scenario 3}

Sarah is a 7-year-old girl struggling with maths due to her introverted nature. She often finds it challenging to ask questions in class, leading to a growing gap in her understanding. Sarah's school has recently introduced EduBot, an AI-powered social robot designed to assist children with their learning needs. This scenario explores how EduBot transforms Sarah's educational experience and the broader implications of its widespread adoption over the next two decades.

In the first few weeks of EduBot's introduction, Sarah is hesitant to interact with the robot. However, she soon discovers that EduBot is approachable and patient, providing clear explanations for her math problems without judgment. EduBot's child-like appearance and friendly voice make it easy for Sarah to communicate her doubts.

EduBot's interactive tutoring sessions help Sarah understand complex math concepts through engaging activities and personalised feedback. The robot's ability to detect her emotional state ensures that Sarah receives encouragement and support when she feels frustrated or anxious. As a result, Sarah's confidence in her math abilities begins to grow, and she starts participating more actively in class.

EduBot quickly becomes a valuable tool for teachers. By analysing student performance data, the robot identifies areas where students like Sarah need additional support. Teachers use this information to tailor their lessons and provide targeted interventions. EduBot also facilitates group activities, promoting peer-to-peer learning and collaboration among students.

Over time, EduBot's presence in the classroom becomes normalised. Students view the robot as a helpful companion rather than a novelty. Teachers appreciate the robot's ability to handle routine tasks, such as grading and attendance, allowing them to focus more on personalised instruction and mentorship.

\subsection{Scenario 4}

Sarah wakes up to the gentle sound of her favourite music playing softly through her bedroom speakers. She stretches and says, ``Good morning, HomeAssist." The system responds with a cheerful tone, ``Good morning, Sarah! The weather today is sunny with a high of 75 degrees. You have a doctor’s appointment at 10 AM and a call with your sister at 3 PM. Would you like to review your schedule?"

She nods, and the system provides a detailed rundown of her day. Afterwards, it offers, ``Would you like to start your morning routine now?" Sarah agrees, and HomeAssist guides her to the bathroom, reminding her of the location of her toothbrush and other essentials.

Breakfast is Sarah’s next goal. She heads to the kitchen where the Smart Kitchen Assistant is ready to help. ``What would you like to cook today, Sarah?" it asks. ``How about some scrambled eggs and toast?" she replies. The assistant provides step-by-step voice instructions, alerting her when the pan is heated and when to flip the eggs. Sensors in the kitchen monitor the stove and other appliances, ensuring they are used safely. If the assistant detects any safety issue, like a burner left on, it promptly alerts Sarah and offers to turn it off automatically.

\clearpage
\section{HRI Value Compass on paper}
\label{app:canvas}

The latest paper version of the HRI Value Compass is available at \url{https://doi.org/10.5281/zenodo.14002841}.
Note that the font is small as these are supposed to be printed on A4 or A3 paper.

\begin{figure*}[htb]
    \centering
    \includegraphics[height=\linewidth,angle=90]{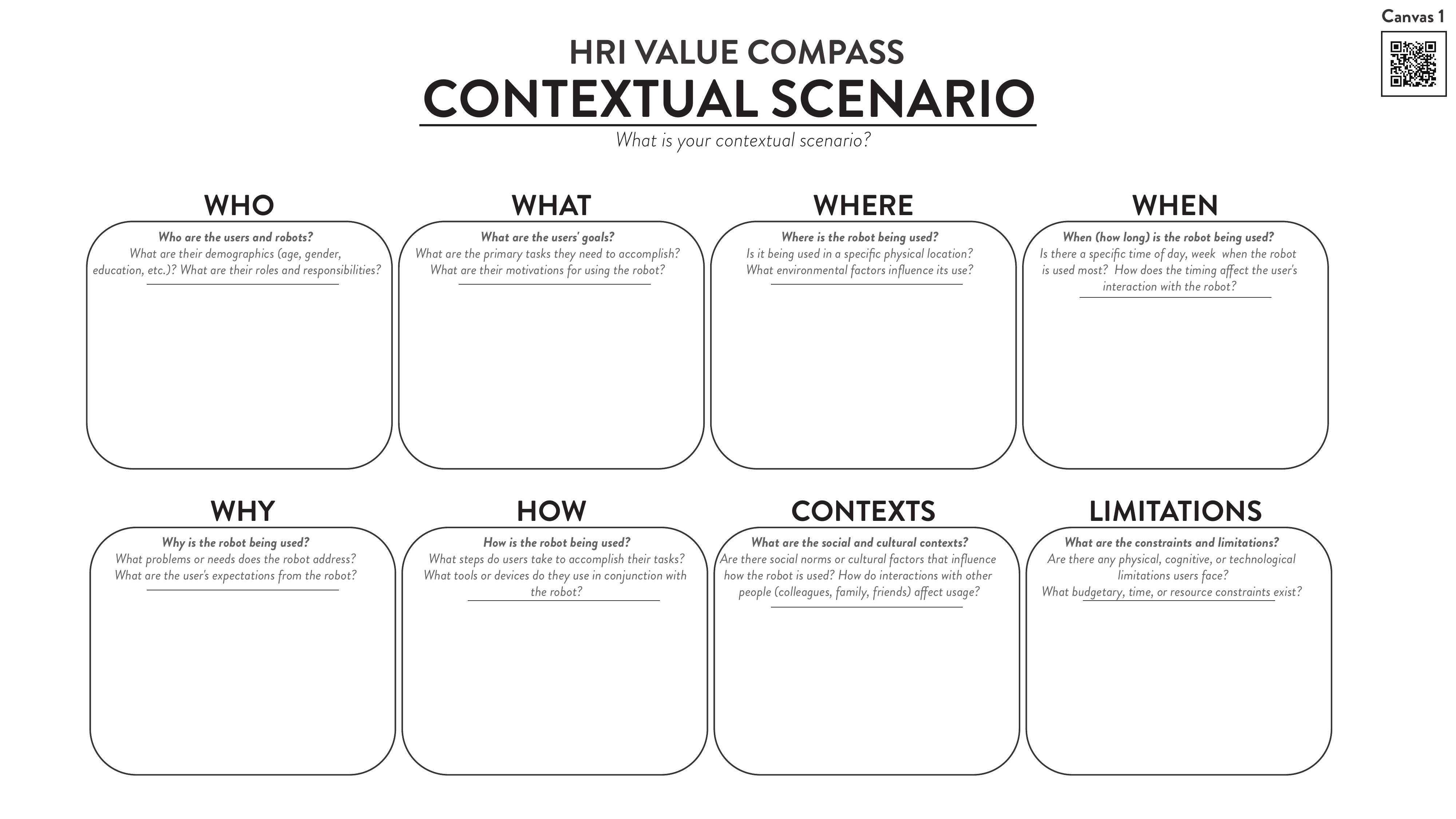}
    \caption{HRI Value Compass paper-version for eliciting values and ethical concerns: contextual information.}
    \label{fig:canvas1}
\end{figure*}

\begin{figure*}[htb]
    \centering
    \includegraphics[height=\linewidth,angle=90]{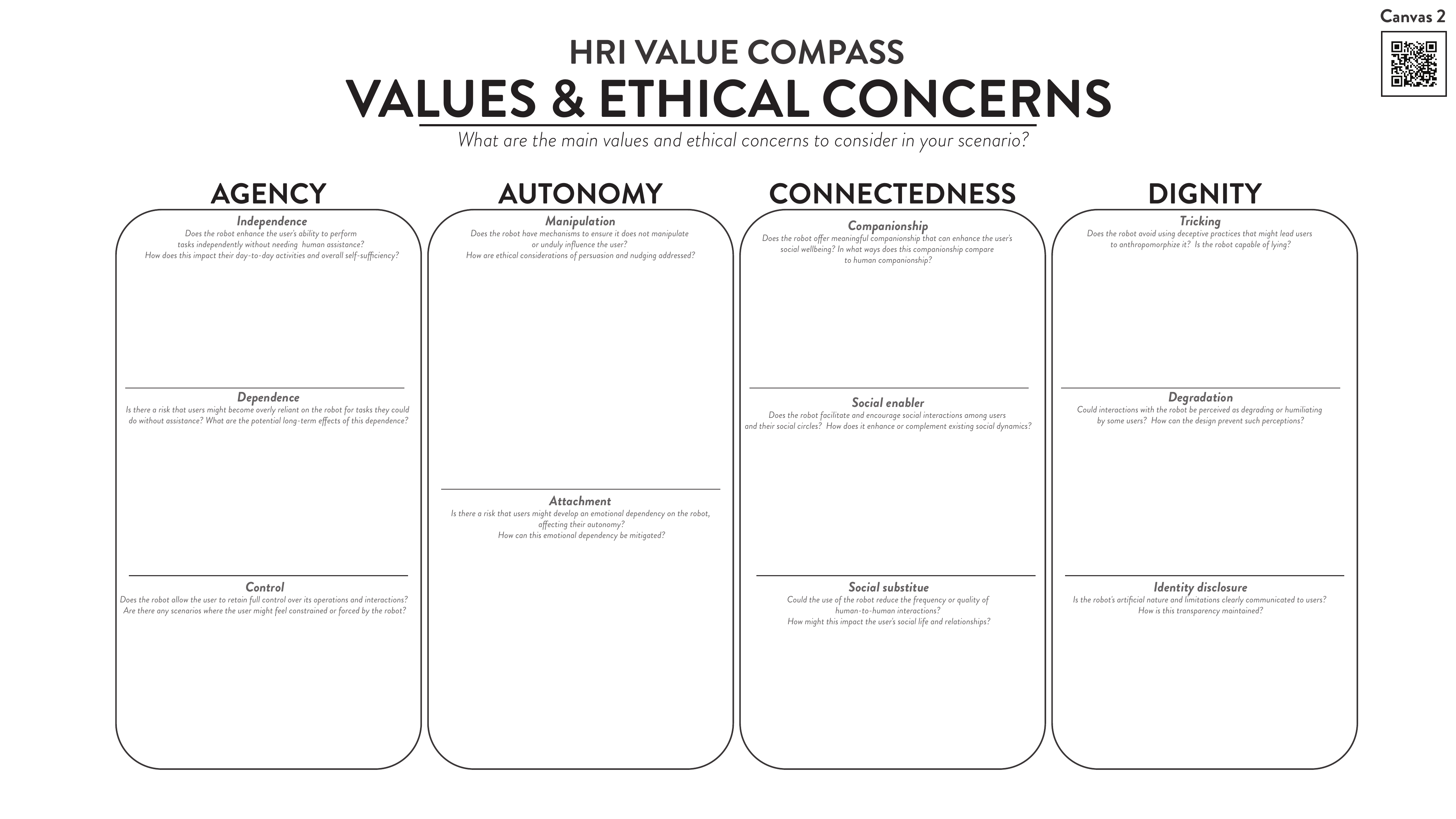}
    \caption{HRI Value Compass paper-version for eliciting values and ethical concerns (1).}
    \label{fig:canvas2}
\end{figure*}

\begin{figure*}[htb]
    \centering
    \includegraphics[height=\linewidth,angle=90]{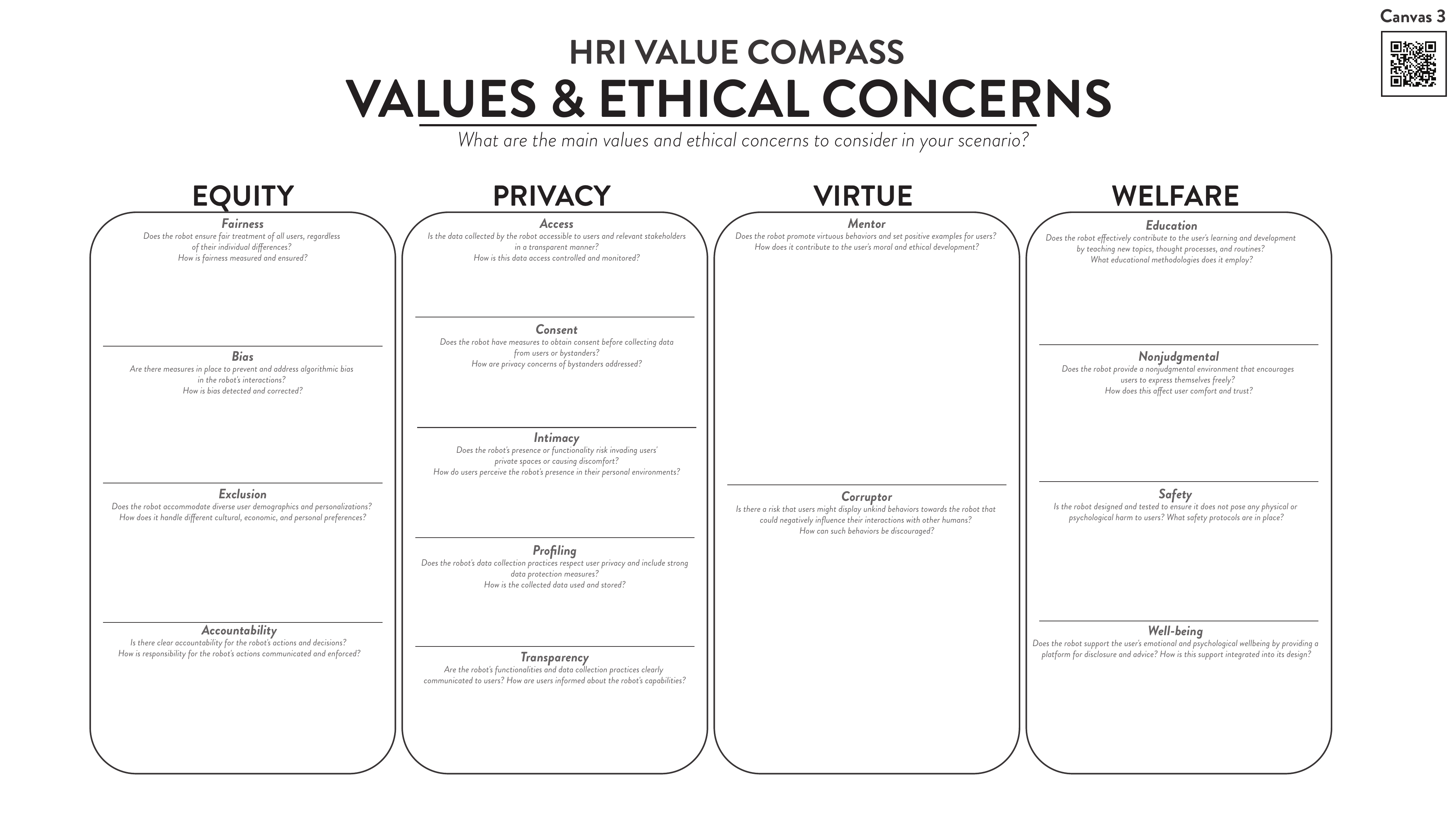}
    \caption{HRI Value Compass paper-version for eliciting values and ethical concerns (2).}
    \label{fig:canvas3}
\end{figure*}

\clearpage
\section{Results from Phase 2: codes and themes}
\label{app:phase2-results}

We first automatically transcribed the recordings of the focus group and we then manually checked them.
Using the framework method for qualitative analysis \cite{braun2012thematic} (i.e., using the values identified in our scoping review as main themes and the topics as main codes), we summarise the focus group findings as follows. In bold we highlight the values that emerged, and in italic the corresponding topics,

\subsection{Activity 1: scenario analysis}

Participants identified various values from the sci-fi HRI scenarios presented to them, without being influenced by any preliminary results from our scoping review that we used as codes to frame our analysis.

Under the umbrella value of \textbf{empowerment}, \textit{data access} emerged as a significant concern in scenarios 1 and 3. Specifically, P4 highlighted the importance of preserving data protection for children in scenario 3 and P1 pointed out that disclaiming upfront who is going to access the data collected in scenario 1 was of extreme importance. 
Analogously, \textit{manipulation} has been identified as a source of concern for scenarios 2,3 and 4. 
P2 highlighted the complexity of scenario 1, which involves children, and argued that children should be meaningfully consulted about their values. P2 also questioned the appropriate ``level of paternalism" to apply in such a complex scenario.
P4 raised concerns about the potential influence of the robot on children's behaviour, particularly in terms of ``\textit{nudging} and manipulation". 
P6 pointed out that in scenario 2, the child, Luke, might not have the capacity to ``understand or consent to the use of the robot", leaving the decision-making responsibility to the caregiver. She also emphasised that, particularly in scenario 3, it is crucial to establish clear guidelines on the extent to which students or learners can exercise agency in deciding the types and amounts of interactions they will engage in.

Participants identified \textbf{privacy} as a key concern in scenarios 1 and 4. P1 questioned whether the Kim in scenario 1 is ``truly able to consent to data collection and use," given her limited ``understanding of the technology." P3 noted that privacy was the main issue in scenario 4, as it takes place in a ``private environment where legislation is likely more relaxed". 
Also in terms of \textit{intimacy}, P1 suggested that there could be a ``panopticon effect" in scenario 1 on the Kim's behaviour. Assuming that Kim ``is aware she is always being observed" and that various stakeholders could access her data, P1 questioned whether Kim might ``self-police" her behavior, acting in ways she believes would be acceptable to potential observers. P1 also argued that the child could ``effectively lose access to her private life" due to the robot's constant presence and monitoring.
The problem of \textit{profiling} has emerged in all four scenarios. In scenario 1, P1 questioned whether the data collection is always in Kim's ``best interest." For instance, P1 wondered if the data could be shared with Kim's insurance company or sold to third parties.
In scenario 2, P1 noted that these concerns are ``even more pronounced" because Luke is a child. He argued that when forming a relationship with the robot, Luke may not be able to fully ``consent to how his data is being used," suggesting that such decisions should be left to his parents or guardians.
In scenario 3, P5 introduced the concept of ``datification" (i.e., reducing people to data), emphasising that designers and researchers should avoid collecting data on children's abilities in ways that could lead to discrimination. Instead, they stressed the importance of preventing the datification of childhood altogether.
In scenario 4, participants pointed out that a home environment may be more sensitive because it is not restricted to a medical setting where stricter rules may apply, increasing the potential risks and harms. They specifically mentioned that in a domestic context, data-capturing systems are particularly vulnerable to ``misuse by domestic abusers."

Regarding \textbf{welfare}, P1 and P3 emphasised that in scenario 1, \textit{safety} related to Kim's physical \textit{well-being} in health contexts is extremely important. P3 also suggested that there might be a ``slight restriction" on Kim's autonomy when the robot provides comfort to her during moments of distress. P5 also highlighted the importance of medical values, especially in scenario 1, for ensuring beneficial interactions for the target user.

Participants raised the issue of \textbf{equality} (referred to as equity), which could lead to both positive and negative outcomes. 
In scenario 1, P2 pointed out that while the robot observes Kim's behaviour and responds accordingly, it is unclear how the robot ensures that the values it displays truly reflect Kim's desires and interests. For scenario 2, P1 noted that the robot will ``inevitably make certain assumptions" leading to \textit{exclusion} based on its programming about what constitutes ``normal" behaviour, and these assumptions could have problematic effects. This highlights that social behaviour may vary depending on different contextual factors, such as culture.
P1 expressed several concerns about potential algorithmic bias in EduBot in scenario 3. For example, he noted that the robot might struggle to accurately identify emotional states in children with different cognitive abilities or socioeconomic backgrounds, or it might ``treat different kinds of students differently" in undesirable ways, such as reinforcing gender stereotypes.

Regarding \textbf{connectedness}, P1 described the robot as a \textit{social enabler} in scenario 1, as it helps Kim stay in touch with her family. However, he also warned that over the long term, the robot could become a \textit{social substitute}, weakening Kim's connections with her family. For example, family members might feel less inclined to check on Kim, knowing that the robot serves as a companion.
In scenario 2, P2 also expressed concerns about the robot acting as a social substitute in educational settings. He questioned how teachers could effectively interact with and provide feedback to students if all teaching activities are delegated to the robot.

P3 pointed out that \textbf{autonomy} could be risky, particularly in scenario 2, where the robot is designed to increase the child's independence. However, this approach may imply that autism needs to be ``corrected," which could be problematic.
In scenario 4, participants noted that the robot might lead to increased \textit{attachment} and ``isolation from other people." Conversely, P1 suggested that a robot like HomeAssist could help alleviate household burdens, potentially reducing gender disparity by taking on tasks like cleaning and cooking, which might lighten the load for women and encourage a more balanced distribution of responsibilities.
P2 and P5 also questioned whether relying heavily on the robot could lead to a deterioration of various skills, such as cooking or reasoning (\textit{dependency}). They wondered if this ``deskilling" might be offset by the development of new, possibly superior, skills gained through interaction with the robot. 
On a positive note, P6 highlighted that robots, as depicted in scenario 4, have the potential to enhance the user's sense of ``self-determination."

In scenario 1, the value of \textbf{honour} was discussed in relation to \textit{agency}. P2 questioned how design choices for the robot, such as expressive eyes and a gentle voice, might ``mislead Kim in some significant way." P2 also explored what values might be implied by these design features.

Finally, the value of \textbf{truth} was addressed in the context of scenario 1. P1 argued that Kim should be informed about how the data collected by the robot is used, and that the robot should be \textit{transparent} about this.

\subsection{Activity 2: values revision}
After completing Activity 1, participants proceeded to Activity 2. In this activity, one of the researchers presented the values identified in the scoping review. Participants were then asked to reflect on the scenarios they had just discussed, considering these values, and to revise these values and topics based on their expertise.

Findings from this activity revealed the following pitfalls in the values identified by the scoping review. 

P4 noted that \textbf{autonomy} and \textbf{empowerment} are closely related, suggesting that the concept of empowerment should be integrated into autonomy, which he defined as the ``capacity of a person to make decisions independently and rationally." P4 emphasised that while independence from others is important, it is not the same as autonomy, which involves making decisions without external impositions. He argued that manipulation should also be considered under autonomy because it impacts decision-making.
P1 agreed that the value of autonomy needed clarification. He proposed distinguishing between two types of autonomy: rational decision-making and agency, which involves having options and the ability to act on them \cite{prunkl2022human}.
P5 pointed out that autonomy should also be considered in relation to humans. It is possible to undermine a person's autonomy even when it is independent of other humans. In scenario 4, for example, the individual does not rely on other humans for support or decision-making.
P4, P1, and P5 agreed that honour should be redefined as \textbf{dignity} and should not include agency. P4 also suggested that manipulation should be categorised under autonomy rather than empowerment and recommended adding the concept of \textbf{nudging}, which was previously missing.
Regarding \textbf{privacy}, P1 pointed out that having a robot present at all times could undermine autonomy and dignity, as constant surveillance might lead to persistent worry. He also clarified that \textbf{fairness} encompasses not just equal treatment but also addresses algorithmic bias.
P5 emphasised that privacy considerations can have positive aspects when data is collected in compliance with established regulations, as such data can be valuable for improving technology.
P3 emphasised the critical importance of the value and \textbf{ethics of care}, suggesting it should be explicitly recognised rather than being subsumed under welfare. He proposed rewording equality as \textbf{equity}.
P1 felt that truth was not the best term to use, arguing that accountability is more relevant. He also recommended adding \textbf{security} to address concerns about potential control and misuse of the system and data. Additionally, P1 highlighted the importance of \textbf{power}, particularly in terms of who makes decisions, such as a single corporation controlling all robots.
Lastly, P4 argued that kindness should be considered a virtue rather than a value and suggested removing it from the list.
P5 emphasised the critical importance of contextualising values, noting that they are situation-dependent and must be prioritised based on the specific circumstances.
P6 commented on the overall categorisation of values and topics, suggesting that some topics could themselves be considered values, as noted in the ethics of AI literature \cite{hagendorff2020ethics}. 

\subsection{Activity 3: contextual information and relationship with values}

Findings from Activity 3 highlighted the relationship between contextual information and moral values. All participants acknowledged that each piece of contextual information could be connected to various moral values to some extent. However, during the discussion, they focused primarily on the connections they found most relevant.

Participants believed that \textbf{empowerment} is strongly related to the ``who" (specifically the role of the robot) and the ``what" (the users' goals). P1 suggested that this moral principle depends on whether the robot is replacing a human role, enhancing connections, or enabling new activities. P3 emphasised that to truly empower users, the robot’s primary task should not be to replace humans but to support them. For example, in math learning scenarios, the robot’s goal should focus on teaching skills rather than the subject itself.
P5 raised the issue of situations where an elderly individual has written advanced directives, and the robot is responsible for calling for assistance. The question is how to handle these autonomous decisions made by the robot.  
P6  highlighted that the users are not necessarily able to fully understand what their needs are and what they want. 
She also pointed out that the users' desires may differ from what the caregiver believes is best for them, which could create tension when defining values.
\textbf{Privacy} was considered as closely related to the ``where," ``how," and ``what contexts" in which the robot is used. P3 emphasised that using a robot at home is ``completely different" from using it in a public space. P1 pointed out that how the robot is used is also crucial; for instance, if the robot is connected to other devices, it could impact the flow of data in and out of the robot. P4 highlighted the importance of the social and contextual environment in which the robot operates. For example, if a person ``does not want others to know they are using the robot," privacy should be maintained in all use cases. P5 suggested that the current legal scenario is important to take into account when complying with regulatory frameworks.
For \textbf{welfare}, participants believed it was linked to the ``where," ``when," and ``why" the robot is used. P1 emphasised that safety must be ensured in all environments, particularly when robot navigation is involved. P4 highlighted that welfare is crucial when there are constraints on when the robot should be used; for example, ensuring that a robot is not used by a child at night. P4 also noted that the ultimate purpose of using a robot should be to promote welfare by making ``life easier" and ``people happier".
Participants discussed \textbf{equality}, also referred to as equity, as being related to the ``who," ``where," ``how," and ``what constraints" when using the robot. P1 noted that equity is connected to demographics and that the environment ("where") significantly impacts the robot's performance. For example, if a robot overheats, its location might affect how easily it can be repaired. P3 added that the way a robot is used can limit access; for instance, if it requires another device to operate, some people may not be able to use it. Regarding constraints, P1 expressed concerns about manipulation, deception, and other ``cognitive limitations" that could make people more susceptible to a robot's behaviour. 
P5 thought it was important to distinguish between different geographical regions and cultures to understand the values that predominate in a given context. For example, in the EU, there's a strong focus on prioritising the rights of individuals, which influences how technology, like autonomous robots, is used and developed.
P6 highlighted that for the definition of ``who" designers should wonder if they are choosing the right kind of people by considering at-risk groups, vulnerability, and the inclusion or exclusion of stakeholders. 

\textbf{Autonomy} was considered to be related to the ``who" and ``what" of using a robot. P4 emphasised that when designing a robot, the focus should be on enhancing the user's autonomy, not the robot's. It is crucial to protect vulnerable populations, such as children and the elderly, and support their independence. P4 viewed the robot as a tool, an ``enabler" that helps users achieve autonomy, and suggested implementing age-related restrictions on its use. P1 highlighted the importance of the robot assisting in decision-making that aligns with the person's best interests and enabling them to do things they otherwise could not.
Participants viewed \textbf{honour} as related to cultural dignity and social norms concerning how robots should behave with the users they interact with.
\textbf{Truth} was discussed in relation to the ``where" and ``why" a robot is used. P1 noted that accountability and control over a robot can vary depending on the scenario. Similarly, P4 emphasised the importance of being transparent about the reasons for using the robot.

\end{document}